# FROM TOKENS TO AGENTS: A RESEARCHER'S GUIDE TO UNDERSTANDING LARGE LANGUAGE MODELS

**Author**: Daniele Barolo 0009-0003-1793-5929

**Abstract**: Researchers face a critical choice: how to use—or not use—large language models in their work. Using them well requires understanding the mechanisms that shape what LLMs can and cannot do. This chapter makes LLMs comprehensible without requiring technical expertise, breaking down six essential components: pre-training data, tokenization and embeddings, transformer architecture, probabilistic generation, alignment, and agentic capabilities. Each component is analyzed through both technical foundations and research implications, identifying specific affordances and limitations. Rather than prescriptive guidance, the chapter develops a framework for reasoning critically about whether and how LLMs fit specific research needs, finally illustrated through an extended case study on simulating social media dynamics with LLM-based agents.

**Keywords**: Large Language Models, Transformer Architecture, AI Literacy, LLMs for Research, Model Context Protocol, LLM Agents.

## Introduction: Opening the Black Box

*Between Hype and Skepticism*

Every researcher today faces a choice that will shape the validity of their work: how to use—or not use—Large Language Models (LLMs). The tools powering ChatGPT, Gemini, and similar systems have become ubiquitous in writing, coding, and assisting with a great variety of tasks. It is no surprise that they have been increasingly used in research as well (Liang et al., 2024), across every phase of a study. From the ideation of research questions (Yang et al., 2024), to literature review (Y. Lee et al., 2025), to assisting with coding and data analysis (Miao, Davis, Pritchard, et al., 2025), to serving as methodology for the research itself - e.g. data annotation (Törnberg, 2023), synthetic data generation (Argyle et al., 2023), simulations (Park et al., 2023) - to the final writing phase (Z. Lin, 2025). There have even been attempts at replacing the whole science production process (C. Lu et al., 2024).

This widespread adoption pushes researchers into one of two problematic camps: those who *anthropomorphize* these systems and trust LLM outputs too readily, and *black-box skeptics* who dismiss them as too complex or unreliable for serious use.

Both stances often stem from a mismatch between how these systems are commonly *perceived* and what they *actually are* (Maeda & Quan-Haase, 2024) – and thus what they can do. On one hand, the anthropomorphizing camp falls victim to a dominant framing that presents LLMs as conversational partners that "understand" and "reason". Indeed, trustworthy (Y. Liu et al., 2024), intelligent, and friendly AIs (Ibrahim, Akbulut, et al., 2025) are positioned as advanced and marketable[1]. On the other hand, the black-box skeptics may equally lack an understanding of the fundamental structure of those models and thus their capabilities when used critically.

Here we pursue the harder but wiser middle path. We ask ourselves: if we strip away the conversational interface we have grown used to, what are LLMs? And, most importantly, how can answering this fundamental question help us use them appropriately?

*Chapter Outline*

This chapter does exactly that: it opens the black box as much as possible or at least helps us get familiar with the main components that constitute the box. By very concretely understanding what, for example, is a context window, the parameter size of a model, or the difference between a base and fine-tuned model, we will be able to maintain a more critical approach and better understand the rationale behind using LLMs well for research.

Every section will discuss one of **six key components** that we have identified to help make sense of how LLMs work (see Figure 1). While these categories are somewhat arbitrary—in reality, these elements are deeply intertwined and cannot be neatly separated—breaking them down this way makes the complexity much more approachable.

---

[1]Companies designing these interfaces have strong incentives to create this perception. For example, OpenAI, to introduce one of their most recent models GPT-5, defined it as a "smartest" model featuring "built-in thinking that puts expert-level intelligence in everyone's hands". https://openai.com/index/introducing-gpt-5/ (last accessed 10.12.2025).

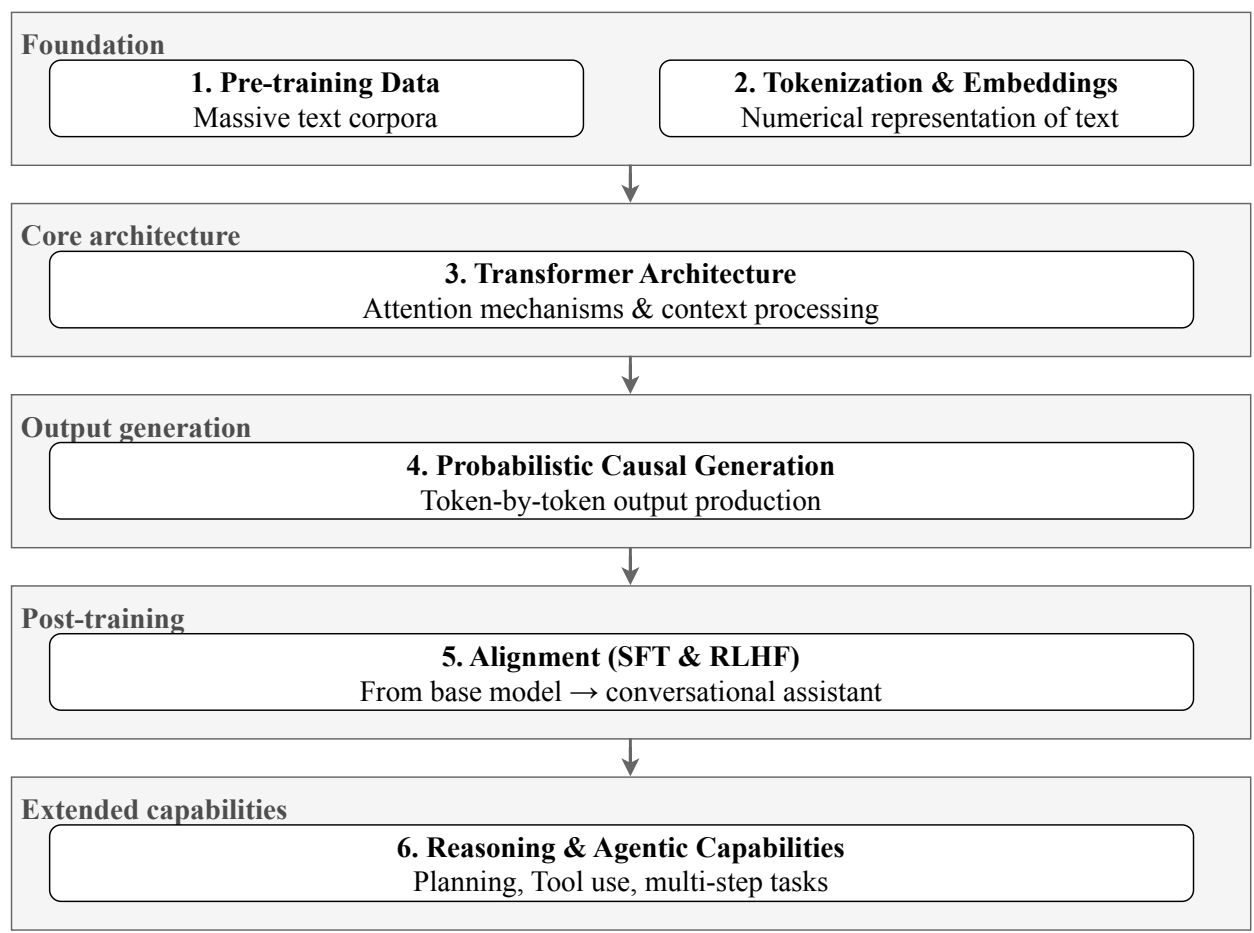

*Figure 1.Six key components of Large Language Models explored in this chapter. SFT = Supervised Fine-Tuning; RLHF = Reinforcement Learning from Human Feedback (see Section 6).*

*Implications for Research*

For each dimension, we will examine both its technical foundations and its practical *implications for research*. Each section follows the same structure: first explaining the dimension itself, then examining its research implications in a box like the following.

> *Affordances*: Each dimension unlocks specific capabilities, practical affordances that create new research opportunities.
>
> *Limitations*: However, each dimension also introduces limitations through built-in restrictions, common failure modes, or subtle risks in research design.

*Box 0. Research Implications Structure*

Each section also includes **'Deep Dive'** references for technical depth (primarily chapters from Jurafsky & Martin's excellent book *Speech and Language Processing* (Jurafsky et al., 2025) along with key papers) and **'Recommended Resources'** for hands-on exploration and visualization.

The chapter concludes with two reference tables: Table 1 compiles all curated resources, and Table 2 synthesizes the affordances and limitations across all six components. Beyond these dimension-by-dimension breakdowns, Box 7 demonstrates this framework in action through an extended case study on simulating social media dynamics. Rather than providing prescriptive answers, it shows *how to reason critically* about whether and how LLMs fit specific research needs.

## 1. Pre-Training Data: The Foundation of What Models Can Do[2]

*What Is In There*

Let us start with something very fundamental: the model's "knowledge" is bounded by its **training data**. LLMs have been trained on more text than any human could read in multiple lifetimes. Yet, they have not *understood* a single word of it in the way we understand this sentence (Suzgun et al., 2025).

What they have done is something perhaps more interesting and definitely more useful for certain tasks: they have mapped the statistical distribution of human language—though not all languages equally, but primarily those well-represented in their training data. This means that the patterns they can reproduce are fundamentally limited to the patterns they have encountered. Since overrepresented patterns in training data become overrepresented in outputs, while underrepresented ones remain rare, it is thus very crucial to understand what this data looks like.

The very large text corpora used for pre-training is essentially the whole web or at least a massive portion of it. One of the main sources comes from **Common Crawl**[3], a nonprofit organization that regularly crawls and archives billions of web pages from across the internet. Think of it as an automated snapshot of everything publicly available online. Model developers then process this raw data into cleaner versions, such as the Colossal Clean Crawled Corpus (C4) (Dodge et al., 2021; Raffel et al., 2023).

This web foundation is then augmented with more carefully **curated sources**: open code repositories like GitHub[4], digitized books (e.g. Project Gutenberg[5]), academic papers from sources

---

like Semantic Scholar (Lo et al., 2020) or arXiv[6], and encyclopedic content like Wikipedia. The final scale of the training corpus is enormous. Modern LLMs train on trillions of tokens (roughly, chunks of words—we will address the difference between tokens and words in Section 2) over hundreds of days on massive GPU clusters[7].

Due to this lengthy training process, models have a **knowledge cutoff**. This is the date when data collection stopped (and likely the training of the model started). For example, GPT-5 —which was released on August 7, 2025— has a September 30, 2024 knowledge cutoff, meaning it had no direct access during training to any information after that date.

*How do I Know What Data a Model was Trained On?*

Given how crucial training data is we might expect we would have complete information about what goes into each model. Unfortunately, we do not[8]. What we know depends entirely on what companies or research teams choose to disclose. This roughly creates three categories: proprietary models, open-weight models, and truly open-source models.

**Proprietary** models are the least transparent. Companies like OpenAI, Anthropic, or Google can independently decide what to reveal about their training data. Usually, they share basic information like the knowledge cutoff, number of tokens (size of the corpus), or main data sources. But the detailed composition usually remains private—for example, which corpora were used or how they were pre-processed.

**Open-weight models** occupy a middle ground. Companies release the complete model weights (the learned parameters—something that will become clearer in Section 3), making them accessible on platforms like Hugging Face[9]. Examples include LLaMA 3 models from Meta (Grattafiori et al., 2024), Gemma 3 models from Google (Team G., et al., 2025) or DeepSeek R1 (DeepSeek-AI et al., 2025). While this accessibility fosters reproducibility in research settings, as we can re-run the exact model with identical parameters, open weights does not mean open data.

---

[6] ArXiv is an open-access research-sharing platform, specialized in STEM fields, that by now hosts more than two million scholarly articles. See more at https://arxiv.org/ (last accessed 10.12.2025).

[7] Note: Many current foundation models are multimodal, trained on text, images, audio, and video, enabling them to process and generate across these formats (Carolan et al., 2024). However, to keep the scope manageable, this chapter focuses on text-only models.

[8] However, attempts to estimate and aggregate data on training dataset sizes exist. **Recommended resource**: for an interactive visualization of these metrics over time, see https://epoch.ai/data/ai-models (last accessed 10.12.2025).

[9] The Hugging Face ecosystem is a collaborative platform that hosts and shares open models, datasets, and tools for building and deploying machine learning systems. The full list of available models can be found at: https://huggingface.co/models (last accessed 10.12.2025).

There are (few) projects that try to truly develop **open-source models** that release everything: data, code, training logs, checkpoints. These are rare but extremely valuable for studying, for instance, how training data ultimately shapes model behavior. The **OLMo 2** family of models (from AI2) exemplifies this (OLMo et al., 2025). The same research team, to address the transparency gap, also built and released **Dolma** (Soldaini et al., 2024), the full pretraining dataset used for the first family of OLMo models (Figure 2).

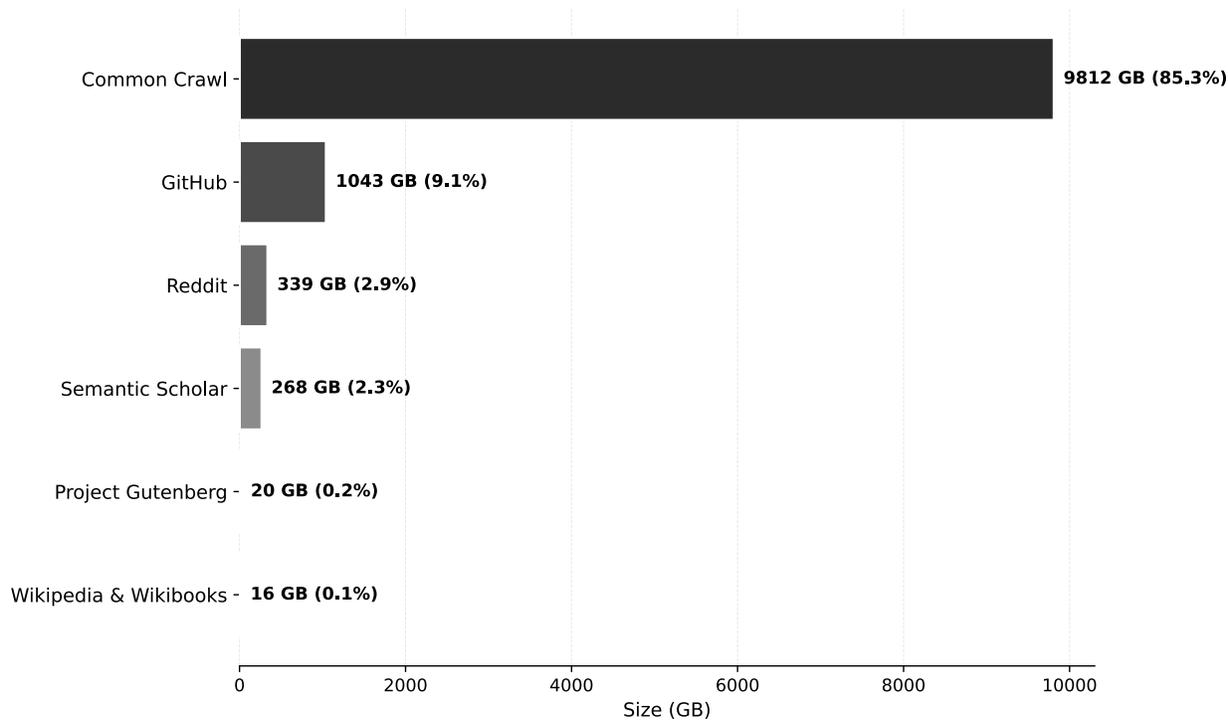

*Figure 2. Composition of the Dolma pretraining corpus by data source. The dataset comprises three trillion tokens from diverse sources, with Common Crawl web data constituting 85.3% of the total. Source data originated from approximately 200 TB of raw text.*

*Affordances*: Pre-training on diverse, massive corpora creates a **broad knowledge base** that enables models to handle varied research tasks (from literature synthesis to text classification) without domain-specific retraining (Brown et al., 2020; Radford et al., 2019). This same diversity provides **zero-shot capabilities**, allowing models to perform tasks they were not explicitly trained for by drawing on the very broad patterns across their training data (T. Wang et al., 2022). Finally, for **open-source models with transparent data**, researchers gain **predictable bias profiles**: knowing what is overrepresented (and what is missing) allows to anticipate failure modes and interpret outputs more critically (Laurençon et al., 2023).

*Limitations*: Models are **bounded by training patterns**. If the research focuses on niche scientific domains or non-English contexts underrepresented in training data, unreliable outputs or confident *hallucinations* are highly expected (Ji et al., 2023; Rawte et al., 2023). Every model has a **knowledge cutoff**, meaning research on recent policy changes, contemporary innovation trends, or current scientific developments will be incomplete or outdated without additional information (Lewis et al., 2020) or external tools (Schick et al., 2023). **Representational bias** systematically skews outputs toward domains highly present in training data, making current models less reliable for studying marginalized communities, Global South contexts, or alternative innovation pathways (Gallegos et al., 2024). Finally, **limited generalization** means models struggle when the research application requires reasoning about scenarios substantially different from their training examples (e.g., predicting novel policy outcomes or analyzing *what-if* scenarios) (A. Hosseini et al., 2022).

*Box 1. Research Implications: Pre-training Data*

## 2. Tokenization and Embeddings: How Models "See" Language[10]

*Tokens Instead of Words*

Until recently, large language models could not tell how many *r*'s are in the word "strawberry". Instead, they would confidently give the wrong answer[11]. Why? Because the model never actually saw the word "strawberry" as a sequence of letters s-t-r-a-w-b-e-r-r-y in the first place. Models don't process language the way we experience it. When you read this sentence, you see words, letters, punctuation. When instead a model processes this sentence, it "sees" a sequence of tokens: chunks that might be words, might be parts of words, might be individual characters.

---

[10] **Deep Dive**: "Speech and Language Processing" §2 (Jurafsky et al., 2025) and for a deeper introduction with interactive visualizations to the static embeddings, see the excellent Lena Voita's "NLP Course" (Voita, 2020) https://lena-voita.github.io/nlp_course/word_embeddings.html (last accessed 10.12.2025).
[11] https://community.openai.com/t/incorrect-count-of-r-characters-in-the-word-strawberry/829618 (last accessed 01.12.2025)

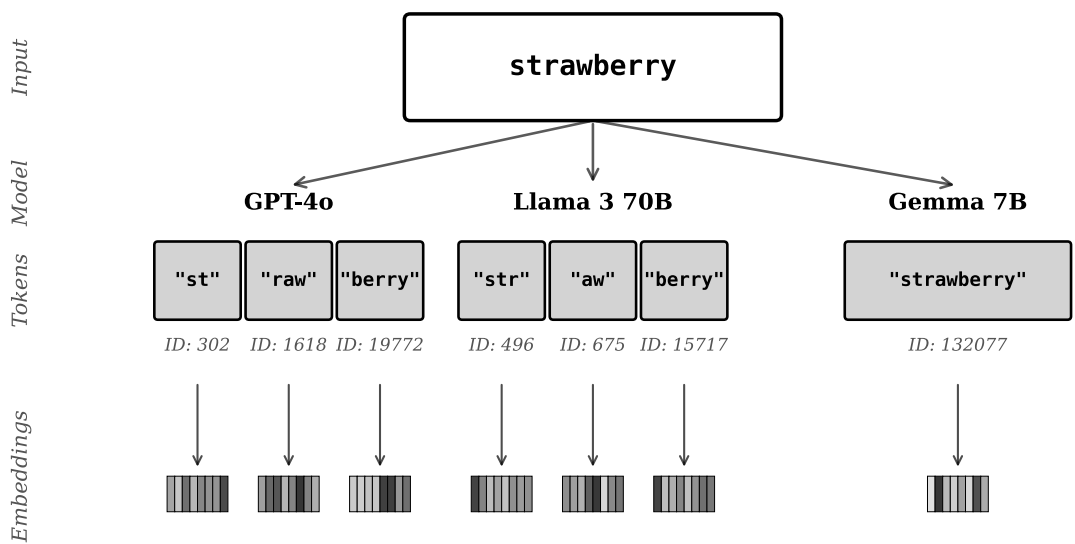

*Figure 3. Tokenization differences across models for the word "strawberry". The token IDs shown are the actual values used by each model. The reader can experiment with different text and models using Tiktokenizer (see footnote 14).*

Since computers process numbers, not text, language needs to be converted into something computational. Words seem like the natural unit—each word mapped to one number. However, new terms emerge constantly[12] and no vocabulary list, however large, can capture every possible word. This approach would require either an infinitely expanding vocabulary or result in countless "unknown" terms the model cannot process. The solution modern language models use is called **Byte-Pair Encoding**, or BPE (Sennrich et al., 2016b), an algorithm that, instead of whole words, learns **tokens** (chunk of words[13]) based on their frequency. This results in a vocabulary list that typically contains 50 to 200 thousand tokens. As for the pre-training data, models also differ in how they *tokenize* (break words into their corresponding tokens) words in a sentence[14] (Figure 3.1).

*From Tokens to Vectors*

Tokens are mapped to IDs in a vocabulary, numbers like 5847 or 21093. They do not mean anything by themselves. How does a model learn that token 5847 (suppose "cat") and token 8392 ("dog") are related, while token 5847 ("cat") and token 3021 ("democracy") are not? Each token

---

[12] For example, terms such as "COVID-19," technical jargon, proper names etc. This is formalized by Herdan's Law (Ross, 1960) or Heaps' Law (Heaps, 1978). The vocabulary size (number of unique words) grows roughly as the square root of text length. In simple words, the larger the corpus, the more new word types appear; meaning no finite vocabulary can ever capture all possible words.

[13] In the final vocabulary, most tokens are actually complete words, but approximately 30-40% represent sub word units.

[14] **Recommended resource**: for an interactive visualization of how models tokenize a given sentence differently try the Tiktokenizer: https://tiktokenizer.vercel.app/ (last accessed 22.12.2025).

ID is then mapped to an **embedding**, a vector of hundreds to thousands of numbers that encodes the semantic content of the token. The embedding relies on the **distributional hypothesis** from linguistics, often summarized as "you shall know a word by the company it keeps" (Firth, 1957; Harris, 1954; Joos, 1950). Tokens that appear in similar contexts are assigned similar vectors, allowing semantic relationships to emerge from patterns of co-occurrence.

**Word2Vec** (Mikolov, Chen, et al., 2013; Mikolov, Sutskever, et al., 2013) is a well-established method that operationalized this principle directly[15]. The algorithm (called skip-gram with negative sampling, or SGNS for short) trains on a simple prediction task: given a token, predict its neighboring tokens (or vice versa) from the sentences it appears in. Through millions of such predictions, tokens appearing in similar contexts naturally acquire similar vector representations. From this emerges something remarkable: the very geometry of the embedding space captures semantic relationships (see Figure 3.2). The now-famous example is vector arithmetic like king - man + woman ≈ queen (Allen & Hospedales, 2019) where directional relationships in the vector space correspond to conceptual relationships (gender in this example)[16]. Think of embeddings as organizing words along multiple dimensions simultaneously: one dimension might capture gender (separating "king" from "queen"), but another might capture status (grouping "king" and "queen" together, away from "peasant")[17].

---

[15] Another established method is GloVe (Pennington et al., 2014)—it achieves similar results through a different mechanism, learning embeddings from global co-occurrence statistics rather than local prediction windows.

[16] **Recommended resource**: The user can visually explore the vector space using the Embedding Projector tool: https://projector.tensorflow.org/ (last accessed 14.12.2025).

[17] This spatial organization of vectors emerges naturally from how these words are actually used in language: "king" and "queen" appear in similar contexts (royal contexts), while "king" and "man" share gendered contexts, allowing the model to discover these relationships through co-occurrence patterns.

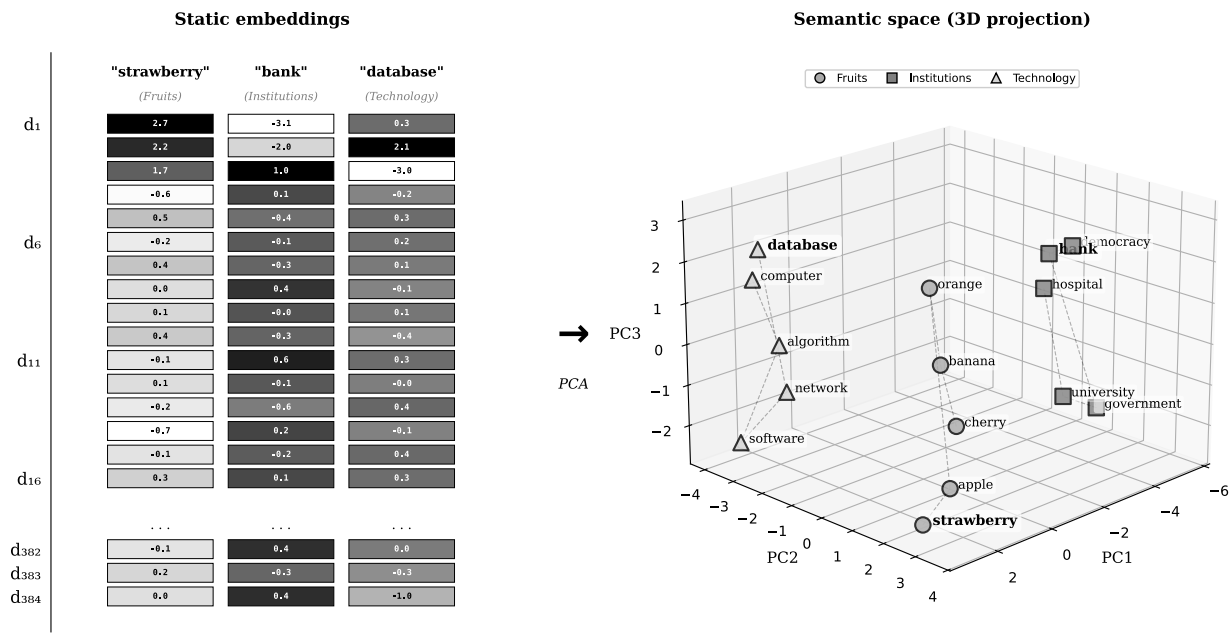

*Figure 4. Visualization of static word embeddings (Word2Vec) in semantic space. Left: embedding vectors for three words across multiple dimensions. Right: 3D projection via Principal Component Analysis (Hotelling, 1933) showing semantic clustering—fruits (circles), institutions (squares), and technology terms (triangles).*

However, there is a crucial difference between those early static embeddings and modern LLMs. Word2Vec gives each word one fixed embedding, which creates a fundamental problem: polysemous words like "bank" receive identical representations whether referring to a river *bank* or financial *bank*, despite having entirely different meanings. Current language models do something more sophisticated: they use **dynamic embeddings**. Unlike Word2Vec's fixed representations, these embeddings change based on the surrounding words in a sentence. The initial token embedding is just the starting point. As the full sentence flows through the transformer's many layers (discussed in Section 3), each token's representation gets continuously updated based on its context[18].

---

[18] For example, when processing "I went to the river bank" versus "I deposited money at the bank," an LLM starts with the same initial embedding for "bank" in both cases, but by the end of processing, the two instances of "bank" have acquired very different representations—one influenced by "river," the other by "money" and "deposited."

*Affordances*: Tokenization and embeddings enable **semantic text operations** like measuring conceptual similarity between documents, clustering research by topic, or retrieving related literature landscapes (Devlin et al., 2019; Mikolov, Chen, et al., 2013), which can be useful for systematic reviews or mapping research (Alchokr et al., 2023). Sub word tokenization handles diverse text including typos, neologisms, and mixed-language content by breaking unfamiliar terms into known components, providing some **robustness to textual variation** (Kudo & Richardson, 2018; Sennrich et al., 2016a). These representations ultimately have enabled **efficient processing at scale**, allowing models to handle massive text corpora that would be impractical for human analysis (Peng et al., 2021; Shibayama et al., 2021).

*Limitations*: On the other hand, sub word fragmentation means rare words, non-English text, and specialized terminology get split into multiple tokens, **degrading performance on rare/specialized terms** such as technical jargon, proper nouns, institutional names, and emerging concepts (Chai et al., 2024). This could be problematic when analyzing scientific literature with novel terminology or conducting multilingual research. Thus, **multilingual performance gaps** emerge once more because non-English languages require more tokens per semantic unit, making models less efficient and reliable for research in these contexts research (Kanjirangat et al., 2025; Petrov et al., 2023). **Token-level biases** create systematic output patterns where frequent tokens get sampled more readily (Martinez et al., 2024), affecting for example citation generation (plausible but fabricated references), simulated responses (convergence on high-frequency options), and any research task where what matters conceptually doesn't align with what was common in training data.

*Box 2. Research Implications: Tokenization and Embeddings*

## 3. Transformer Architecture and Attention Mechanism[19].

Now we address another side of the black box, arguably the most important one. We have seen that LLMs use dynamic, context-dependent embeddings. But how does this actually happen?

The core mechanism is that as text flows through the LLM structure, each token does not just carry its initial embedding forward, instead actively examines surrounding tokens, gathering relevant context and updating its representation accordingly. This happens through a specific mechanism called the **attention mechanism**, which is applied repeatedly across many layers, progressively refining what each token "means" within its context. These final, context-rich representations are then used to predict the next token in the sequence (we will discuss this further in Section 4).

*The Attention Mechanism: Learning What Matters*

---

[19] **Deep Dive**: "Speech and Language Processing" §7 and §8 (Jurafsky et al., 2025) as well as "NLP Course" (Voita, 2020) https://lena-voita.github.io/nlp_course/seq2seq_and_attention.html (last accessed 10.12.2025).

The shift from static to dynamic embeddings relies on what is known as attention (or *self-attention*) mechanism. The attention lets the model dynamically compute which parts of the input are most relevant to understanding each token in context. While the formal description of the self-attention mechanism would deserve a chapter of its own, we will try to show step-by-step the intuition of how the mechanism works when processing a given sentence.

The first step is **positional encodings**. Since attention-based models have no inherent sense of word order (unlike previous sequential models that process words left-to-right[20]), an additional embedding that tells the model where each token sits in the sequence is added to the static token embeddings. This combined representation (token embedding + positional encoding) becomes the input to the first transformer layer.

Next, for each token in the sequence, the model creates three different vectors by multiplying its input embedding by three *attention weight matrices* learned during pre-training. Think of these as three different "views" of the same token, created through three different learned transformations (see Figure 5 for a visual overview of this process).

- A **query** vector (representing what information this token needs)

- A **key** vector (what this token offers as information)

- A **value** vector (the actual information this token provides)

To figure out how much attention one token (let us call it token A) should *attend to* another token (token B), the model computes the dot product (a way to measure similarity between two vectors) between A's query vector and B's key vector. This produces a single number called an **attention score** (or weight). A high attention score means the vectors align well; a low attention score means they do not. Finally, the model uses these attention weights to compute each token's output representation. This is a weighted sum (using the attention score values) of all the value vectors in the sequence. If token A assigned high attention weight to token B, then B's value vector contributes heavily to A's updated representation.

[20] For more information about the history of Language Models before the GPT-like one, refer to "Speech and Language Processing" §3,4, and 13 (Jurafsky et al., 2025).

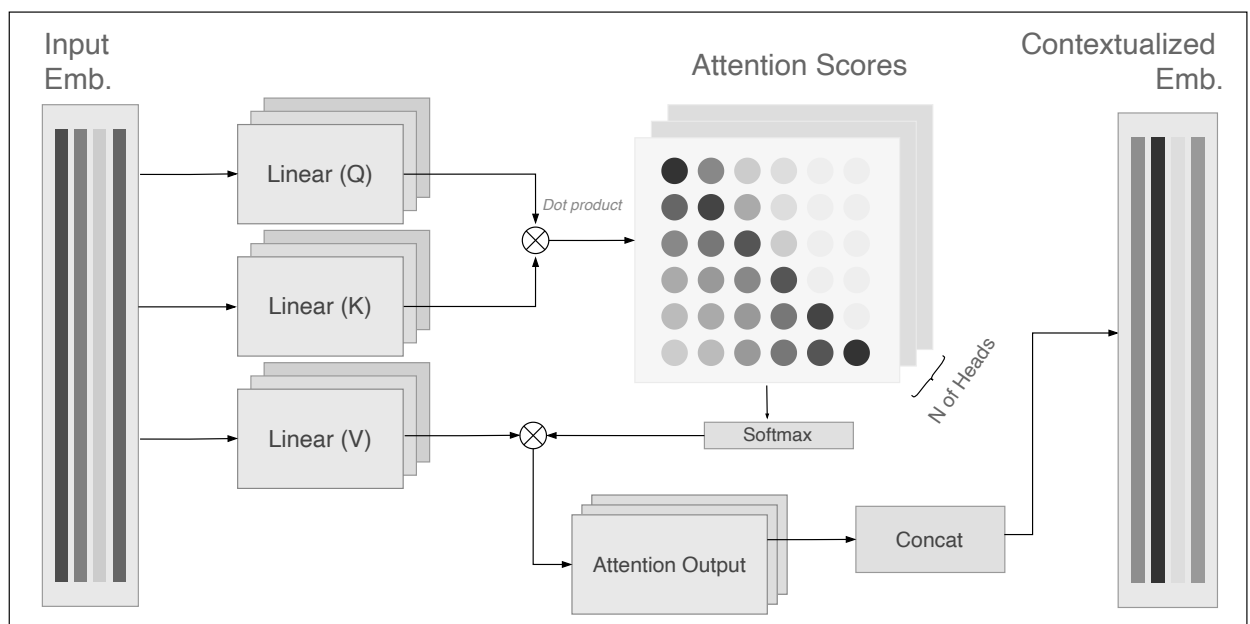

*Figure 5. Overview of the self-attention mechanism. Input embeddings are transformed through linear projections (Q, K, V). Attention scores (grid) determine token relationships. Weighted sum produces causalized embeddings. For interactive exploration of this process, see the Transformer Explainer tool (footnote 24).*

So far, we have described only one head of the self-attention layer, but the mechanism actually consists of several heads (e.g., 12 in GPT-2-small). Hence the term **multi-head attention**[21]. Each head uses its own set of query/key/value projections, focusing on different patterns like grammar, meaning, or long-range links. Research shows some specialized heads consistently attend to adjacent positions (positional patterns), while others track syntactic relationships like subject-verb agreement across long distances (Voita et al., 2019). Note that no one explicitly programmed which relationships to learn, rather the model learned how to discover these patterns automatically during training.

*The Transformer Architecture*

The multi-head attention mechanism we just described is the heart of the transformer (the architecture at the core of the current LLMs), but it does not work alone. Instead, a complete **transformer block** (or layer) consists of three key components:

1. Multi-head attention layer (what we just explained).

2. Feedforward networks (FFN), also called *multilayer perceptrons* (MLP).

3. Residual connections.

---

[21] Note: more current models, like Llama 3 (Grattafiori et al., 2024), use **Grouped Query Attention** (GQA), that is sharing key/value heads across query groups for efficiency while retaining multi-head benefits.

Within a transformer block, after multi-head attention processes the tokens, the output passes through a **feed-forward network**—a simple neural network that transforms each token's representation independently. The FFN combines linear projections (which expand the representation to a higher dimension) with non-linear activation functions, then projects back to the original size. This combination of linear and non-linear operations allows the network to learn the complex patterns in language data[22].

Instead of completely replacing an input's representation at each step, **residual connections** add the output of each component (attention or feed-forward) back to the original input. This way, even after passing through dozens of layers, the original input information remains accessible.

Finally, multiple transformer blocks (each with multi-head attention) stack on top of each other[23]. These blocks form a **residual stream** where information flows upward: each block's output is added back to the ongoing representation rather than replacing it, allowing information to accumulate across layers[24] (Figure 6).

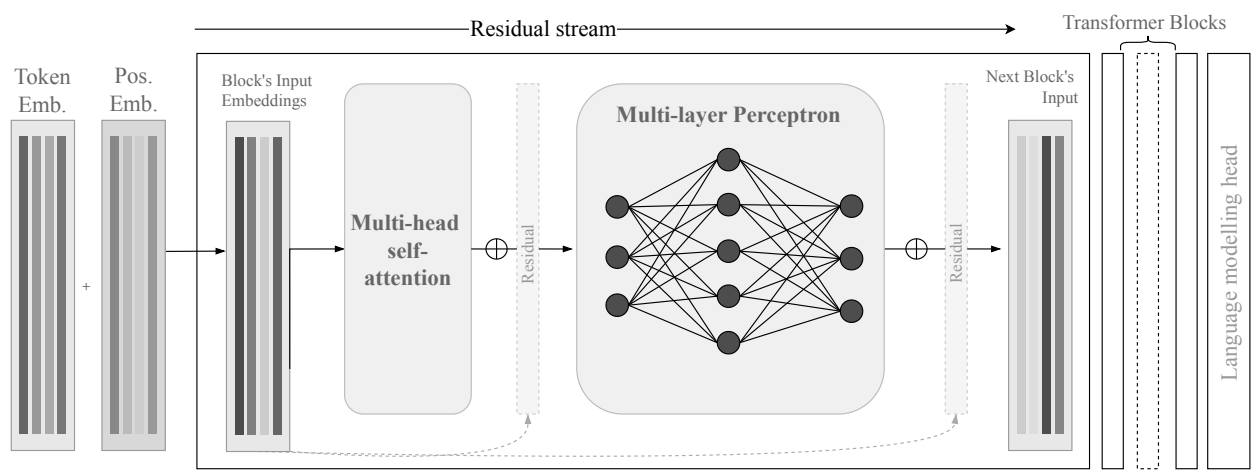

*Figure 6. Schematic representation of the transformer architecture showing stacked blocks and residual stream. Token embeddings flow through multiple blocks, each applying multi-head self-attention and feed-forward networks with residual connections. For interactive exploration of this process, see the Transformer Explainer tool (footnote 24).*

As representations flow upward through these stacked blocks, token vectors progressively refine their meaning. Early layers learn simple patterns like positional relationships, while deeper layers (closer to the output) capture long-distance dependencies and higher-order semantic implications (Ferrando & Voita, 2024).

---

[22] Interestingly, these feed-forward layers contain about two-thirds of the model's total parameters, while the other third lives in those pre-learned attention weight matrices discussed above.

[23] Current LLMs' architectures consist of 12 to more than 100 blocks (e.g., LLaMA 3.1 405B has 126 blocks (Grattafiori et al., 2024)).

[24] **Recommended resource**: Visual step-by-step walkthroughs of a full LLM forward pass include bbycroft.net/llm (last accessed 05.12.2025) and the Transformer Explainer https://poloclub.github.io/transformer-explainer/ (last accessed 10.12.2025).

The model's "knowledge" lives in its weights. Millions or billions of numerical parameters encode linguistic patterns learned during training. Research on neural **scaling laws** reveals a consistent pattern: larger training datasets, more compute power, and higher parameter counts all improve model performance (Kaplan et al., 2020). Crucially, when models scale up along these dimensions in a balanced way, they don't just get incrementally better, they begin displaying qualitatively new capabilities on certain tasks, often referred to as **emergent abilities** (Wei et al., 2022). The key insight for us that that size matters: larger models usually can do more tasks, and do them better.

*Encoder-Decoder Types*

The original transformer architecture (Vaswani et al., 2017) combined an encoder and a decoder. The **encoder** part takes a sequence of tokens as input and produces a contextualized vector representation for each one. It achieves this through self-attention layers that let all tokens examine each other simultaneously in both directions. The **decoder** part instead generates output autoregressively (one token at a time, left-to-right) using only the prior part of the sequence (see Figure 7).

Current language models typically use three architectural variants optimized for different tasks.

**Encoder-only** models like BERT (Devlin et al., 2019) are trained by *masking* out words and learning to predict them from surrounding context on both sides. These are not generative models. Instead, they are used for tasks like text classification, named entity recognition, or sentiment analysis, where the need is to analyze the meaning of a text.

**Encoder-decoder** models like T5(Raffel et al., 2023) were originally designed for translation and summarization, where input and output are distinct sequences. While decoder-only models have largely replaced them for general-purpose tasks[25], encoder-decoder architectures can still be advantageous for specific structured transformations where clear input-output separation matters.

---

[25] Thanks to their general-purpose abilities and simple prompt-response interface, decoder-only models (such as the GPT models) have become the most accessible architecture for both research and real-world applications.

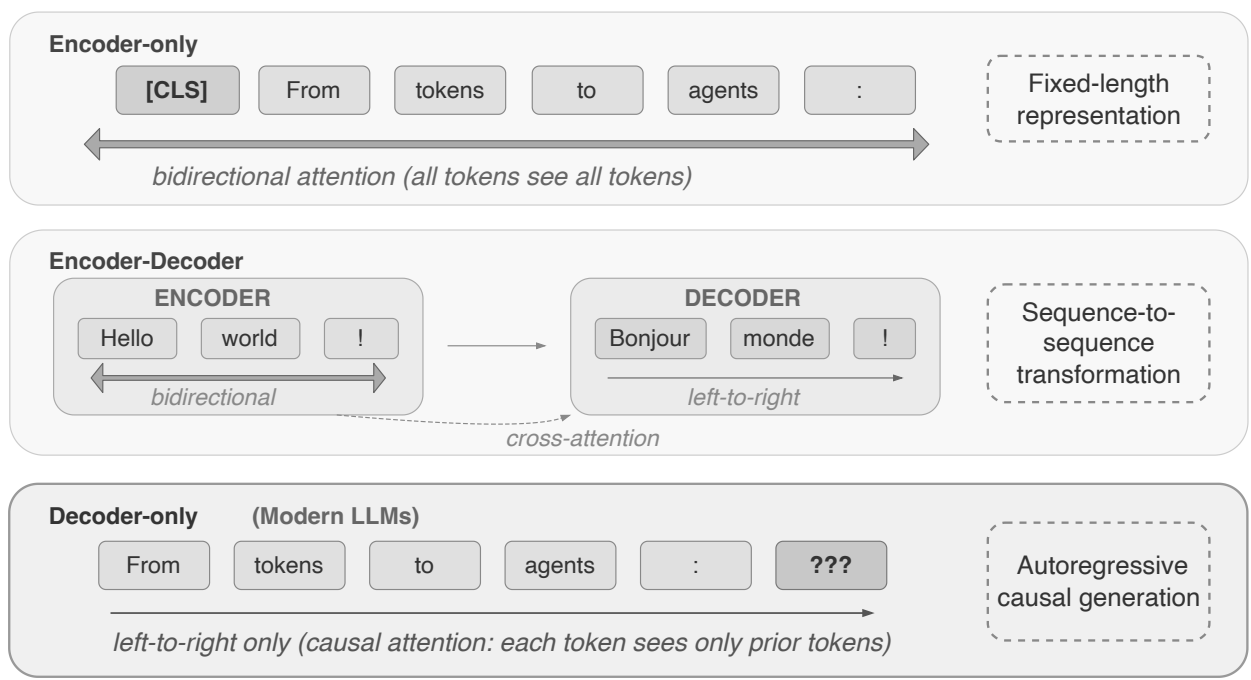

*Figure 7. Three transformer architecture variants. Encoder-only models allow all tokens to attend to each other bidirectionally; sentence-level representations (from <CLS> token or averaged embeddings) enable classification tasks like sentiment analysis (top). Encoder-decoder: bidirectional encoder with causal decoder (middle). Decoder-only: tokens attend only to prior context (bottom).*

Finally, **decoder-only** models are what we mainly refer to as LLMs since the rise of ChatGPT usage (end of 2022) [26]. This class includes models like GPT-4 (OpenAI et al., 2024), DeepSeek R1 (DeepSeek-AI et al., 2025), and Gemini (Comanici et al., 2025). Information flow in decoders goes left-to-right, meaning the model predicts each next word using only prior words (more on this in Section 4). Their autoregressive generation makes them versatile for research tasks requiring open-ended text production such as writing, synthesis, coding assistance, and data generation.

*Context Window and Prompting*

Computing attention between all pairs of tokens is computationally expensive. Specifically, it scales quadratically with sequence length. This ultimately creates a hard limit called the **context window**: the maximum number of tokens a model can process at once[27]. Whatever text is fitted in the context window, which is called the **prompt**, defines everything the model can "see" during inference, and thus what will condition the generation.

---

[26] Note: ChatGPT is the user-friendly app built around a base GPT model (like GPT-5) as it adds extras such as a fixed system prompt for behavior, safety filters to block harmful replies, tools for chat memory etc.

[27] For example, GPT-5 can handle up to 400,000 tokens (272,000 input and 128,000 output) https://platform.openai.com/docs/models/gpt-5 (last accessed 10.12.2025)

This is why prompting matters architecturally[28]. When prompting, we are working within a fixed token budget, which forces decisions about *what* information fits in the context window, such as whether to include task descriptions or a small set of illustrative examples (**few-shot prompting** (Brown et al., 2020)), and *how* that information is ordered[29].

*Current Efficiency Innovations*

While the standard transformer architecture is powerful, it is *dense*, meaning every single parameter is active for every single token. As models scale to hundreds of billions of parameters, this becomes inefficient. Recent models address this through **Mixture of Experts** (MoE) architecture: instead of one large network, the model contains multiple specialized sub-networks (*experts*), with a gating network selecting only the most relevant ones for each token (e.g., 2 out of 8, or 8 out of 256) [30]. Another way to reduce the cost of running these large models is **quantization** (J. Lin et al., 2024; Z. Liu et al., 2024). The strategy is to reduce the numerical precision of model weights and activations (the values computed during inference)[31]. An alternative current solution is **knowledge distillation** (Hinton et al., 2015) where a large, capable "teacher" model is used to train a smaller, faster "student" model. Rather than training from scratch, the student learns to mimic the teacher's outputs and "reasoning" patterns (see Section 6 for reasoning abilities)[32].

Knowing these distinctions helps researchers balance capability and cost: we can often achieve adequate performance with quantized or distilled models while reserving expensive full-scale models for tasks that truly require them.

---

[28] Note: *prompt engineering* is the work of designing natural language instructions to steer large language models toward specific outputs without parameter updates (Brown et al., 2020).

[29] For example, information at the beginning or end of the context often receives more attention than content buried in the middle (N. F. Liu et al., 2024).

[30] A model like DeepSeek-V3, for instance, utilizes dense layers for its initial foundation (layers 1–3) but switches to MoE for layers 4–61 (DeepSeek-AI, Liu, et al., 2025). Model names often encode MoE architecture details: for example, in Qwen3-30B-A3B (https://huggingface.co/Qwen/Qwen3-30B-A3B), "30B" indicates total parameters while "A3B" indicates approximately 3 billion parameters are activated per token.

[31] Training typically uses 32-bit or 16-bit floating-point numbers; quantization compresses these to 8-bit or 4-bit. While this on one hand means a slight accuracy reduction, on the other hand allows to fit larger models into limited GPU memory (VRAM) and significantly accelerate inference.

[32] For example, DeepSeek-R1 used a 800,000-sample dataset generated by its flagship model to fine-tune (see section 5 for details on fine-tuning methods) smaller open-weights models (like Llama or Qwen from 1.5B to 70B parameters). That retained much of the reasoning capability at a fraction of the computational cost (DeepSeek-AI et al., 2025b).

*Affordances***:** The attention mechanism enables models to **select relevant context**, focusing on pertinent parts of input text—connecting pronouns to referents, linking instructions to data, or extracting information from specific document sections (Vaswani et al., 2017). Multi-head attention and stacked layers enable **multi-layer inference**: early layers learn surface features like syntax and formatting, while deeper layers capture semantic relationships and long-distance dependencies (Osama et al., 2019; Tenney et al., 2019), supporting complex inference like detecting implicit meanings or inferring causal relationships not explicitly stated. Large context windows (100K-1M+ tokens) enable **full-document processing** without chunking—entire research papers, policy documents, or transcripts can be analyzed while maintaining coherent understanding across long texts (Reid et al., 2024). Finally, **efficient deployment options** like mixture-of-experts, quantization, and knowledge distillation offer valid and more accessible alternatives to largest models for resource-constrained research settings (Zhu et al., 2024).

*Limitations***:** On the other hand, **context-window constraints** create a hard limit on input size—everything must fit within the context window, and filling it completely degrades performance (X. Zhang et al., 2024). For complex research tasks requiring extensive context (e.g., simulating multi-turn interactions, analyzing large datasets at once), researchers must strategically decide what to include and exclude (Park et al., 2023). **Position bias** means tokens at the beginning and end of prompts receive more attention than middle content (Guo & Vosoughi, 2025), affecting where critical instructions or data should be placed, particularly important when designing prompts for annotation tasks or structured information extraction (Amor et al., 2024). Finally, **performance-efficiency trade-offs** require balancing model size against resources: model capabilities and knowledge both scale with parameter count, meaning very small models are inadequate for most research tasks (lacking both reasoning ability and world knowledge), while large models are better at complex tasks but demand substantial computational resources (expensive API costs or slower inference). Researchers must navigate this spectrum, often settling for mid-sized or quantized models that balance capability with accessibility (Wan et al., 2024).

*Box 3. Research Implications: Transformer Architecture and Attention Mechanism.*

## 4. Probabilistic Causal Generation: How Models Produce Text[33].

We have seen how models process input through attention and transformers. Now we examine how they produce output. The final layer of the transformer stack, called the **language modeling head**, converts internal representations into text predictions (Figure 8).

*Causal Generation*

Given an input sequence of tokens, a decoder-only model predicts the next token, adds that token to the sequence, then predicts again, one token at a time in a loop. This process is referred to as **causal** (or conditional), as everything that came before, the original prompt plus all previously generated tokens, influences what will come next[34]. The causal generation continues until the model outputs a special **end-of-sequence token** (like `<EOS>`)[35] or reaches a maximum length set by the context window.

---

[33] **Deep Dive**: "Speech and Language Processing" §7 and §10 (Jurafsky et al., 2025) as well as "NLP Course" (Voita, 2020) https://lena-voita.github.io/nlp_course/text_classification.html (last accessed 10.12.2025).

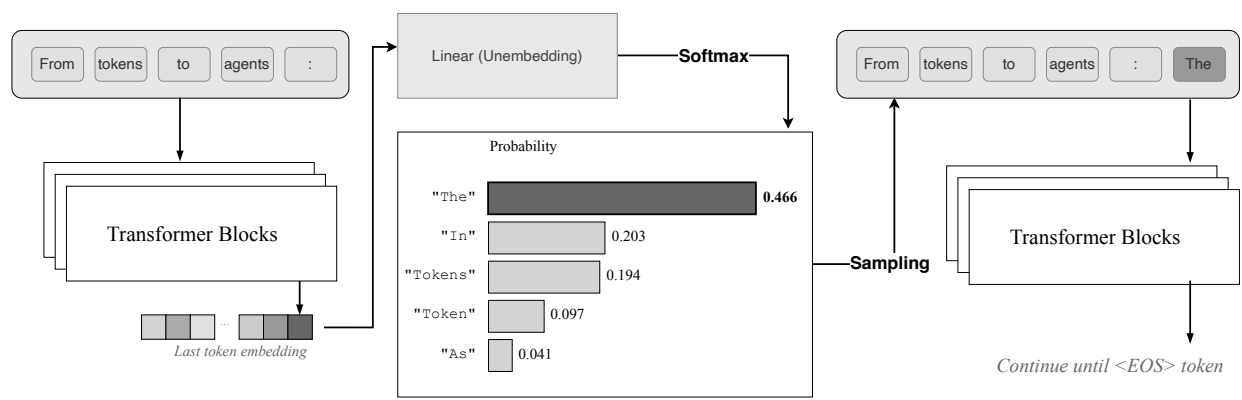

*Figure 8. Overview of the language modeling head in decoder-only models. The input sequence flows through transformer blocks. The language modeling head converts the final token's representation into a probability distribution. Sampled token is appended and fed back until <EOS> or maximum length is reached.*

This is the same process the model used to "learn" language. As we discussed in Section 1, during pretraining the model processes trillions of tokens of text. At each position of a given sequence, all the upcoming tokens of the sequence get masked. Then the model tries to predict the next (correct) token[36]. This strategy is very effective because the text corpus provides automatic supervision, or rather **self-supervision**, and no human-labeled data (which is very costly) is required. Which token gets selected during the causal generation process depends on **sampling**— a process we explore next. The sampled token is compared to the correct token, and the *cross-entropy loss* gets computed (essentially, how "surprised" the model is by the actual next token). The model then updates its parameters to reduce that surprise, gradually aligning with statistically likely linguistic patterns[37].

*Sampling*

---

[34] This process is called **autoregressive** because it uses its own output as input for the next generation step.

[35] During pre-training, `<EOS>` (or similar) token is artificially appended to the end of every training example, teaching the model to predict and output it as a natural "stop" signal when generating text.

[36] Whatever the sampled token is, during training the model gets fed with the correct one (a technique called *teacher forcing*) and needs to predict the next masked token.

[37] This allows the model to learn - for a given sequence of tokens - the probability distribution of tokens over the vocabulary and sample from this distribution. For instance, after the phrase "The study found that," the model might assign 8% probability to "participants," 6% to "women," 5% to "men," 3% to "students," and decreasing probabilities across thousands of other tokens.

After computing probabilities over the vocabulary, the model needs to select one token. This selection process, often referred to as **decoding** or sampling, directly affects output quality, and can be controlled through sampling parameters[38].

The simplest approach is **greedy decoding**: always pick the highest-probability token. However, this produces boring, repetitive text. Plus, it is deterministic, meaning identical inputs always produce identical outputs. Pure **random sampling** solves the repetition problem but produces low-quality outputs. Even though low-probability tokens are individually unlikely, there are so many weird options in the tail of the distribution that they get selected often enough to produce incoherent text.

The solution most LLMs use is **temperature sampling** (Figure 9). Temperature ($\tau$) is a way to control the stochasticity of the output by reshaping the probability distribution before sampling. With low temperature ($\tau < 1$), the model becomes more deterministic, that is, the model amplifies high probability tokens and suppresses unlikely ones, producing more predictable text. A higher temperature ($\tau > 1$) instead flattens the distribution, increasing diversity at the cost of coherence. Finally, at $\tau = 1$, probabilities remain unchanged[39]. However, it is good practice to test the same prompt on different temperature values and pick the one that best fits the purpose, as well as report these sampling parameters alongside model specifications and prompts.

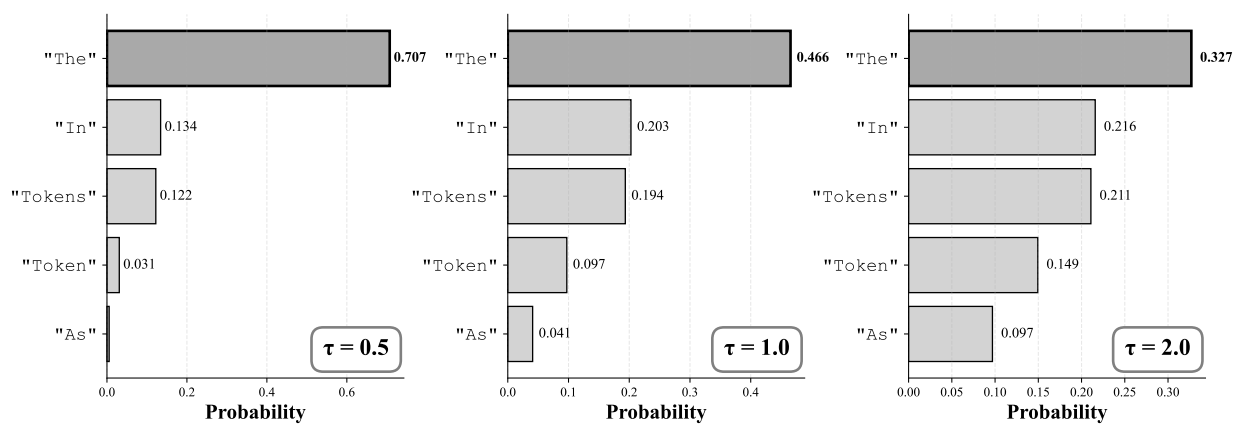

*Figure 9. Effect of temperature on probability distributions during sampling. Note: Shows top 5 token probabilities from Mistral-3B after the prompt "From tokens to agents:", renormalized to sum to 1.0 (a common practice in top-k sampling). Low temperature ($\tau = 0.5$, left) makes the model more deterministic, choosing high-probability tokens. High temperature ($\tau = 2.0$, right) increases randomness by flattening the distribution across alternatives.*

---

[38] **Recommended resource**: Explore decoding and associated probability distributions via the Decoding Visualizer: https://huggingface.co/spaces/agents-course/decoding_visualizer (last accessed 10.12.2025), based on the compact decoder-only LM "SmolLM2-360M-Instruct" developed by Huggingface (Allal et al., 2025).

[39] The rule of thumb for research requiring consistency (e.g., classification, topic extraction, structured outputs) is to use low temperature (0.3-0.7), whereas for exploring diverse possibilities or creative generation, use higher values (1.0-1.5).

Two additional methods control sampling: **top-k** sampling (selecting only from the k most probable tokens) (Fan et al., 2018) and **top-p** (or nucleus) sampling (selecting from tokens comprising the top p% of cumulative probability) (Holtzman et al., 2020). Both methods prevent selecting very improbable tokens while maintaining diversity[40]. Finally, researchers can directly control when **generation stops** by defining custom stopping tokens or sequences (strings that immediately halt generation when produced) or setting a maximum token limit.

*Affordances*: Autoregressive generation enables **natural language abilities** (Brown et al., 2020a; Radford et al., 2019). This makes models useful for e.g., literature synthesis, report drafting, or in general as research tool for any natural language task (Luo et al., 2025; Zhao et al., 2025). Through temperature and sampling parameters researchers have **control over the output diversity**, tuning the stochastic-deterministic trade-off (Holtzman et al., 2019; H. Zhang et al., 2021). Low temperature for reliable classification or annotation tasks where consistency matters (Törnberg, 2024) or higher temperature when generating diverse synthetic data or exploring alternative framings (Agarwal et al., 2024). **Direct probability inspection** provides access to the model's token-level probability distributions (logits), enabling researchers to have an alternative measure of model "confidence" (Jiang et al., 2021; M. Zhang et al., 2024), useful for e.g., estimate response distributions for survey simulation (Cao et al., 2025), or understand which outputs the model considers more or less likely at inference time (Kadavath et al., 2022). Finally, **prompt-guided generation** means carefully structured prompts can steer probability distributions toward desired outputs, what is called *in-context learning* with examples and instructions (Wei et al., 2023).

*Limitations*: Models generate text by sampling from probability distributions, not by retrieving facts from a knowledge base (Mousavi et al., 2025). **Hallucination and factual errors** mean models cannot retrieve specific information like dates, citations, or precise events reliably, instead producing hallucinations —plausible but fabricated information presented with confidence (Huang et al., 2025; Y. Wang et al., 2024). This poses serious challenges for fact-dependent research like bibliometric analysis or literature reviews (Tang et al., 2025). **Lack of logical consistency** means models may contradict earlier statements or fail to maintain coherent reasoning across long generations, requiring researchers to verify logical soundness when inconsistencies would invalidate results (Ghosh et al., 2025). **Bias toward average outputs** occurs because probabilistic generation systematically favors typical training patterns, producing "average" responses while underrepresenting edge cases, rare perspectives, or genuinely diverse behavioral patterns (Sen et al., 2025). This becomes critical when researchers need authentic diversity—for instance, when simulating human behavior with outlier cases or generating synthetic data representing varied populations (A. Liu et al., 2024). Finally, **stochastic output variability** means the same prompt can produce different outputs across runs (Novikova et al., 2025), requiring researchers to assess consistency through repeated trials or direct probability analysis.

---

[40] While these methods can technically be combined with temperature, most researchers adjust just one parameter at a time (usually temperature) as this provides clearer control and easier interpretation of how settings affect outputs.

*Box 4. Research Implications: Probabilistic Causal Generation*

## 5. Alignment: From Base Models to Assistants[41].

Models resulting from pre-training alone complete text—they do not follow instructions (Figure 10). These are called **base models**. Base models are fully functional and publicly available[42], but they fundamentally treat instructions as text to complete rather than commands to execute. Because text prediction was the only training objective, base models also reproduce training data patterns indiscriminately, which often results in misinformation and toxicity. Post-training, organized in two steps, solves both problems, teaching models to distinguish instructions from text and to align outputs with human preferences.

*The Model Learns to Follow Instructions*

The first post-training phase is **supervised fine-tuning** (SFT), which teaches models to recognize and respond to instructions. To do so, SFT uses the same loss function as pre-training (next-token prediction) but this time on high-quality examples of instructions paired with desired responses[43]. Training examples resemble conversations like "Explain photosynthesis in simple terms" → [clear explanation], "Translate this to French" → [accurate translation], "Debug this code" → [corrected code with explanation].

Crucially, SFT **updates all the parameters** of the LLM (see Section 3), reshaping the probability distributions the model uses to predict next tokens by skewing it toward instruction-following patterns. Through SFT, the model learns to recognize task formats, maintain appropriate tone, and produce structured outputs.

---

[41] **Deep Dive**: "Speech and Language Processing" §7 and §9 (Jurafsky et al., 2025).

[42] **Recommended resources**: Often open-weights and open-source models are released together with their base model. For example, the reader can find the base version of the already mentioned Olmo3 7b model here: https://huggingface.co/allenai/Olmo-3-1025-7B and the instruct one here: https://huggingface.co/allenai/Olmo-3-7B-Think-SFT (last accessed 09.12.2025).

[43] Early models like InstructGPT used roughly 13,000 such demonstrations written by human annotators (Ouyang et al., 2022) though modern systems often train on millions of examples combining human-written and synthetic data.

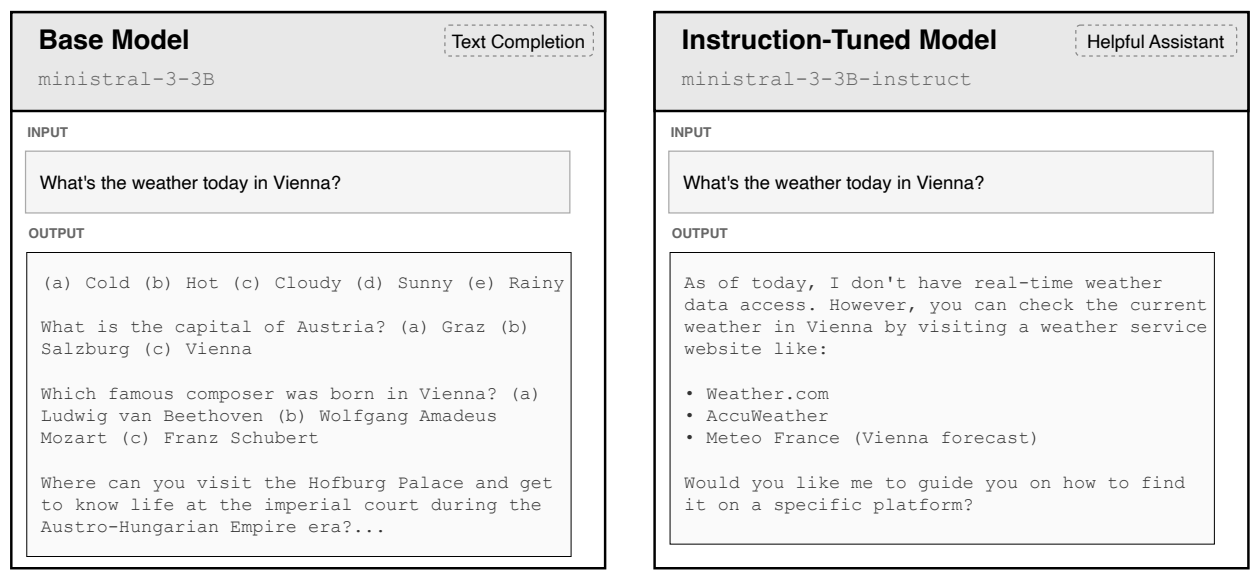

*Figure 10. Comparison of base model versus instruction-tuned model behavior. Note: Both panels show outputs from Ministral-3-3B (December 2024 release) run locally via Ollama with 8-bit quantization. Left panel shows the base model (NikolayKozloff/Ministral-3-3B-Base-2512-Q8_0-GGUF) treating the instruction as text to complete. Right panel shows the instruction-tuned version (mistralai/Ministral-3-3B-Instruct-2512-GGUF) recognizing the input as a question requiring a helpful response.*

### Reinforcement Learning Through Human Feedback

SFT teaches models how to respond to instructions, but not which responses are better than others. The model also needs to learn whether the content is accurate or fabricated, helpful or harmful, and, very importantly, useful for the user. The model learns this through an additional training called **reinforcement learning from human feedback** (RLHF) (Ouyang et al., 2022). The approach can be simplified in three stages.

First, it starts by collecting **preference data** from humans. Human annotators rank multiple model responses to the same prompt. Rather than writing full demonstrations (as for SFT), annotators simply need to compare which is more helpful or accurate. Preference data can also come from implicit web signals, like upvoted comments in Reddit (Askell et al., 2021) or from other AI evaluators rather than humans (H. Lee et al., 2024).

This preference data is then used to train a **reward model**, a separate network that learns to predict human preferences. Given a prompt-response pair, it outputs a numerical score estimating how much humans would prefer that response. This reward model trains on thousands of ranked comparisons, learning patterns in human judgment without needing to generate language itself.

Finally, the policy optimization stage uses the reward model to improve the language model through **reinforcement learning**. Most systems use Proximal Policy Optimization (PPO) (Schulman et al., 2017) which iteratively updates model parameters to maximize reward scores while constraining updates to prevent destroying language capabilities. The language model generates outputs, gets scored, and adjusts its parameters to make high-scoring responses more probable. Recent approaches simplify this pipeline. For example, **Direct Preference Optimization** (DPO) (Rafailov et al., 2023) skips the reward model, directly adjusting the language model's parameters based on preference comparisons. What is most relevant for researchers is that, just as in SFT, RLHF reshapes the probability distributions by making human-preferred answers way more likely than others[44].

*Customized Fine-Tuning*

We have just seen how the core of modern LLMs is developed. However, building a language model from scratch requires resources on a scale few can access, such as vast computing infrastructure, specialized expertise, and costs that can easily stretch into the millions. The good news is that these are general-purpose models and can be used off-the-shelf for research.

But what if we need a model tailored to a specific domain—say, an underrepresented language, a specialized field, or with a particular behavioral characteristic? The most straightforward approach would be **full fine-tuning** one of these pre-trained models. This means continuing the training on new data from a specified domain. Fine-tuning process updates every parameter in the model, whether via supervised learning (as in instruction tuning) or reinforcement learning (as in RLHF).

However, updating billions of parameters remains computationally intensive even when starting from a pre-trained model. Instead, **parameter-efficient fine-tuning** (PEFT) methods have emerged as the practical standard, as they allow researchers to fine-tune only a subset of parameters (Ding et al., 2023). One established method is **LoRA** (Low-Rank Adaptation) (Hu et al., 2021). It works by adding small trainable matrices (called adapters) to the model's layers. These adapters operate in lower dimensions than the full network. This way, it is possible to modify the model's behavior by updating just a few parameters that get activated during inference. The advantage is that many of these **adapters** can be trained on different data sources and swapped in when needed.

---

[44] This reshaping of the model's probability distribution toward human-preferred replies naturally encourages *sycophantic behavior*, since the model is rewarded for acting like a helpful, agreeable assistant (Sharma et al., 2025). Moreover, it also amplifies pre-existing *liberal-leaning political*, reflecting the preferences and moral assumptions of human annotators (Atari et al., 2023; Bang et al., 2024; Fulay et al., 2024; Santurkar et al., 2023).

QLoRA extends this further by quantizing the base model to reduce memory (Dettmers et al., 2023)[45].

*Affordances***:** Post-training transforms base models into **instruction-following** tools that execute commands rather than merely completing text, making them practical for systematic research applications like annotation, classification, or data extraction without extensive prompt engineering (Liyanage et al., 2024; Ziems et al., 2024). **Safety and reduced toxicity** through RLHF filter harmful or toxic outputs that would appear in base models, making aligned models safer for research contexts involving human subjects or public-facing applications (Wuttke et al., 2025). **Accessible customization** via parameter-efficient methods, like LoRA, or fine-tuning of base models, enables researchers to adapt models for specialized domains, underrepresented languages, or specific behavioral characteristics without requiring massive computational resources (Binz et al., 2025). Finally, aligned models excel at **structured output generation**, reliably following format constraints like JSON schemas or tables, valuable for tasks requiring machine-readable outputs (Dagdelen et al., 2024; Y. Lu et al., 2025).

*Limitations***:** Post-training introduces **systematic value biases** by optimizing toward specific annotator preferences and cultural norms. These biases affect outputs when research requires representing diverse populations, non-Western perspectives, or viewpoints outside mainstream values (Arora et al., 2023; Tao et al., 2024; A. Wang et al., 2025)—particularly problematic for cross-cultural studies, simulation of diverse populations, or analyses that rely on LLMs as annotators, where these value biases can be injected directly into the labeled data (Atari et al., 2023; Pieuchon et al., 2025). **Sycophantic behavior** emerges as models optimize for seeming helpful rather than being accurate (Sharma et al., 2025); they agree with user premises, avoid contradictions, and prioritize pleasantness over truth. This becomes critical when research requires adversarial testing, critical analysis, or simulating agents who maintain controversial positions despite social pressure (Chuang et al., 2024). Finally, **overly cautious refusals** (Weidinger et al., 2021) might occur when models pattern-match legitimate research requests to training-time safety concerns, rejecting valid scholarly inquiries about sensitive topics, historical events, or controversial policies that fall within ethical research boundaries.

*Box 5. Research Implications: Alignment.*

## 6. Reasoning and Agentic Capabilities[46]

---

[45] **Recommended resources**: it is possible to fine tune a model, even without coding experience, using this interface https://huggingface.co/docs/autotrain/tasks/llm_finetuning. Otherwise, it is possible to directly implement models fine-tuned by the community and openly released in Huggingface enviroment. For example, the reader can see the full list of all fine-tuned versions (for different purposes) of the model Llama-3.1-8B-Instruct here: https://huggingface.co/models?other=base_model:finetune:meta-llama/Llama-3.1-8B-Instruct (both last accessed 04.12.2025).

[46] **Deep Dive**: Unlike previous sections, no dedicated chapter in (Jurafsky et al., 2025) or lecture in Voita's NLP Course covers these emerging topics comprehensively. This section synthesizes insights from recent research on reasoning capabilities (DeepSeek-AI et al., 2025b), function calling (Schick et al., 2023), and the Model Context Protocol (Hou et al., 2025).

We have examined five core components that define how LLMs fundamentally work: training data, tokenization and embeddings, transformer architecture, probabilistic generation, and post-training alignment. The field evolves rapidly; countless new techniques and architectural innovations emerge constantly, making comprehensive coverage impossible and any such attempt quickly outdated. However, two recent advances have already proven essential enough to warrant attention: reasoning abilities and agentic capabilities[47]. While still evolving, these capabilities are already reshaping what LLMs can do and how researchers can use them

*What Does it Mean for an LLM to Reason?*

The decoder-only language models we have described initially struggled with multi-step reasoning tasks that required breaking problems into sequential steps. This limitation makes sense from an architectural point of view. Transformers process sequences through a fixed number of layers, meaning complex reasoning must compress into relatively few forward passes through the network. When prompted with complex math problems the models often failed because they jumped directly to answers in just one or two tokens, forcing all the "thinking" to happen in those limited computation steps (Wei et al., 2023).

Researchers discovered a surprisingly simple solution called **chain-of-thought prompting**. The technique involves **augmenting few-shot demonstrations** with explicit reasoning steps between the question and answer[48]. This approach proved so effective (Suzgun et al., 2023; Wei et al., 2023) that it inspired a natural next step: what if models learned to reason this way automatically, without requiring carefully crafted prompts?

Current state-of-the-art reasoning models achieve exactly this through reinforcement learning fine-tuning specifically designed for chain-of-thought reasoning (DeepSeek-AI et al., 2025). The training process starts with a small set of high-quality chain-of-thought demonstrations as a "cold start" to initialize the model. **Reinforcement learning with verifiable rewards** then trains the model to consistently produce structured reasoning[49]. The training objective rewards correct answers on reasoning-intensive tasks like mathematics, coding, and logical problems, while maintaining language consistency to keep outputs readable. Some recent models introduce an additional control mechanism called **thinking effort** or reasoning budget (A. Yang et al., 2025), which lets users specify how much computational resources the model should invest in reasoning during inference[50].

---

[47] Indeed, it is enough to look at the description of a recently released OpenAI model, gpt5.1 https://platform.openai.com/docs/models/gpt-5.1 (last accessed 14.12.2025). The model is described as "The best model for coding and **agentic** tasks with configurable **reasoning** effort".

[48] For example, instead of showing "Question: Roger has 5 balls. He buys 2 more. How many does he have? Answer: 7," the demonstration includes "Question: Roger has 5 balls. He buys 2 more. How many does he have? Let's think step by step. Roger started with 5 balls. He buys 2 more, so 5 + 2 = 7. Therefore, the answer is 7."

[49] The model is fine-tuned to include the reasoning trace between opening `<think>` and closing `</think>` tags, while the final response must appear between `<answer>` and `</answer>` tags. During actual use, the thinking tokens get generated by the model and are then parsed away or displayed separately in the interface, allowing users to inspect the reasoning process without affecting the final answer.

*From Next-Token Predictors to Agents*

Reasoning capabilities alone do not solve all limitations. Language models remain constrained by their structure, only able to manipulate probability distributions over tokens. They lack access to current information, cannot compile code, and cannot verify facts against external sources (Mousavi et al., 2025). For example, a model prompted with "What's the weather today in Vienna?" cannot retrieve actual weather data, can only generate hallucinated responses based on training patterns or, if aligned accordingly (see RLHF in Section 5), admit it cannot fulfil the request.

**Function calling** addresses these constraints, an ability that emerged unexpectedly[51] as models grew larger and more capable (Schick et al., 2023). Recognizing this potential, developers fine-tuned the models to make this behavior robust and controllable. These **tool-using models**, when they determine the prompt requires external data, can generate a structured output (typically JSON) specifying a function name and parameters. The function is then executed, and its output is added to the context for the final response generation[52]. Function calling thus transforms the model from a closed system operating on fixed training data into an *open system* that can access external information sources independently[53].

---

[50] Some current models, like GPT-5, do not allow anymore for the temperature parameter, which got replaced by the "thinking effort" parameter. Higher effort produces longer, more detailed reasoning traces for complex problems, while lower effort generates shorter traces for simpler questions. This gives researchers control over the trade-off between response quality and computational cost.

[51] Nobody explicitly programmed it; researchers simply noticed that sufficiently large models could recognize when they needed external data and generate structured requests for it

[52] In our running example (see Figure 10), {"function": "get_weather", "parameters": {"location": "Vienna"}} would be generated by the model. The application wrapping the language model then executes this function call against an external API, receives structured data in return, and appends this data to the model's context, allowing the LLM to generate a final response incorporating the real information. This is how, in a nutshell, systems like ChatGPT retrieve internet information or how Perplexity.ai (https://perplexity.ai, last accessed 10.12.2025) operates.

[53] **Recommended resource**: For an accessible introduction to AI agents, see the open Hugging Face course "Introduction to AI Agents", available at https://huggingface.co/learn/agents-course/en/unit0/introduction (last accessed 14.12.2025).

This function-calling capability enables **LLM-powered agents**, systems that can autonomously combine reasoning, planning, and action to fulfill complex tasks (Yao et al., 2023). An agent might search a database, analyze the results, decide it needs more information, call another tool, evaluate progress, and adjust its strategy until the task is complete.

*Model Context Protocol*

Models can be fine-tuned to use tools. But once a model has this capability, how does it know which specific tools are available? The **Model Context Protocol** (MCP), introduced by Anthropic in November 2024[54], provides a standardized answer.

Instead of hardcoding tool information during training, developers create an **MCP server**—a standardized catalog of available capabilities that any MCP-compatible model can query. This server exposes three types of resources. **Tools** are executable functions the model can call, such as searching academic databases, running statistical analyses, executing code, or querying APIs (like web retrieval). **Resources** are data sources the model can access, including specific PDFs, datasets, supplementary materials from papers, or any structured information repository. Finally, **prompts** are pre-designed workflows containing structured instructions for multi-step tasks[55]. When you connect a model to an MCP server, the model receives this complete catalog of what's available, then it can decide which resources to use based on the task at hand (invoking tools, accessing data, or following structured workflows as needed).

---

[54] Introduced by Anthropic in November 2024 (https://www.anthropic.com/news/model-context-protocol), MCP has rapidly gained adoption, with major providers including OpenAI and Google now supporting the protocol.
[55] For example, a prompt might specify the complete instructions for conducting a systematic literature review, breaking it down into clear steps.

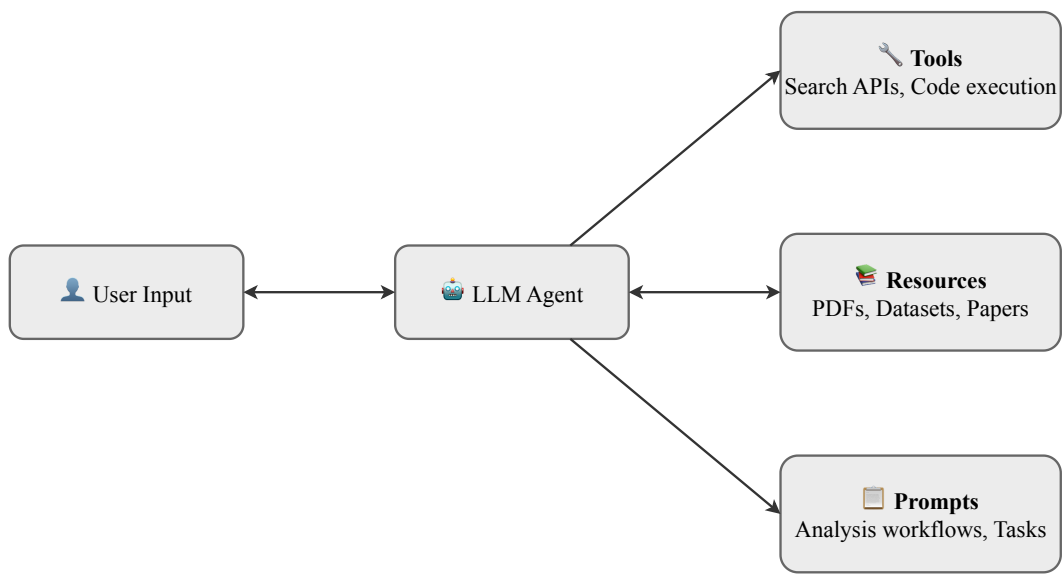

*Figure 11. Overview of Model Context Protocol. An MCP server exposes Tools (executable functions), Resources (data sources), and Prompts (workflows) to LLM agents, enabling dynamic discovery and use of capabilities based on task needs.*

**Source:** Author's own

**Alt Text:** A flowchart showing MCP server architecture. On the left, "User Input" flows to an "LLM Agent" in the center. The LLM Agent connects to an "MCP Server" box on the right, which branches into three rounded rectangles: "Tools" (Search APIs, Code execution), "Resources" (PDFs, Datasets, Papers), and "Prompts" (Analysis workflows, Tasks).

The reason why MCP has become widely adopted is its standardization, which eliminates the need to reconfigure tools for each model or application. By defining a common protocol, developers can build an MCP server once, making it immediately usable by any **MCP-compatible model**[56].

As a practical example of MCP's power, *Paper2Agent* (Miao, Davis, Zhang, et al., 2025) automatically transforms a research paper—parsing its methods, data, and code into executable tools—then deploys them as a dedicated MCP server. Any compatible model can then instantly connect and invoke these contributions via natural language, reproducing analyses or applying them to new data without manual setup. This turns static papers into dynamic, reusable agents, accelerating scientific workflows.

---

[56] **Recommended resources**: There are many openly accessible MCP servers that can be openly used. A detailed list can be found here: https://mcp.so/ or here: https://modelcontextprotocol.io/examples. For more details on MCP lifecycle and construction refer to (Hou et al., 2025).

*Affordances*: Recent advances in **reasoning capability** mean models can now break down complex problems into sequential steps through chain-of-thought approaches, rather than attempting direct answers that often fail. This improves performance on tasks requiring logical deduction, mathematical reasoning, or structured analytical workflows (Lightman et al., 2023). Models have now have **access to current and external information** through function calling and MCP (Hou et al., 2025; Schick et al., 2023), directly addressing the knowledge cutoff and hallucination limitations discussed earlier. For research, this mean model can retrieve real-time data, query external databases, access recent publications, and verify facts against authoritative sources (Mialon et al., 2023). These capabilities ultimately enable **autonomous task execution** where models plan multi-step procedures, execute tools, evaluate intermediate results, and adjust strategies without constant supervision (Yao et al., 2023b). This opens possibilities for complex research applications like comprehensive literature searches, multi-stage data analysis, or systematic evidence synthesis that would require extensive manual coordination otherwise (Glickman & Zhang, 2024). Moreover, the standardization of **reusable and shareable** research tools through MCP means specialized capabilities can be designed once and used across different models and projects, lowering barriers to sophisticated computational methods.

*Limitations*: However, more autonomous agent in research means also **reduced researcher control**, because only the *space* of possible actions is specified (which tools, resources, and prompts are available) but not the exact decision paths the model will take. This complicates reproducibility (Bail, 2024) and makes diagnosing unexpected behaviors more difficult than with fully controlled flows. Additionally, the flexibility also comes with **higher computational expenses** as reasoning budgets, iterative tool calls, large context windows, and multi-step agent loops increase token usage by orders of magnitude, potentially making extensive applications prohibitively expensive for resource-constrained research. Finally, given how recently these capabilities emerged, **best practices are still being established**: empirical evidence about effectiveness for specific research applications is limited, optimal configurations are still under development, and failure modes are being documented (Plaat et al., 2025). Thus, researchers are still required to invest time in trial-and-error exploration.

*Box 6. Research Implications: Reasoning and Agentic Capabilities.*

## Conclusions

*Opening the Black Box*

We opened this chapter by acknowledging a tension: researchers who trust LLM outputs too readily, *anthropomorphizing* these systems as intelligent agents capable of whatever a user requests, and *black-box skeptics* who dismiss them as unreliable for rigorous use in research. Both stances stem from misunderstanding what these systems actually are.

Now, having explored the mechanisms inside the box, we can see a more nuanced picture. Models are neither superhuman intelligences nor completely mysterious artifacts that have innate ability to communicate with us. Instead, LLMs process language fundamentally differently than humans do—through tokenization, statistical patterns, and very complex probability distributions. Yet dismissing them as mere "next-token predictors" oversimplifies their sophistication[57]. Understanding their actual mechanisms—such as how attention flows through layers, how probabilities get shaped by pre-training and alignment, how the context window constrains generation —helps us capture the full complexity at play.

*Six Components, One Framework*

Rather than exhaustively documenting every technical specification (which would quickly become outdated), this chapter prioritized **foundational components** that remain relevant as models evolve. Although these six components are deeply intertwined in practice, examining them separately reveals how each shapes model behavior. **Pre-training data** determines which linguistic and cultural patterns models can reproduce, while **tokenization and embeddings** determine how that material gets processed. **Architecture** sets hard limits on the amount of context that can be processed and determines how information flows through **attention** mechanisms, with deeper layers capturing increasingly abstract relationships. **Causal probabilistic generation** means models reproduce language in the most syntactically *likely* way—this ultimately means that high model confidence does not equal accuracy or factuality. **Alignment** reshapes these probability distributions toward helpful responses, but this process also leads to specific values and model behaviors (e.g., sycophancy (Cheng et al., 2025; Sharma et al., 2025)). Finally, **reasoning and agentic capabilities** enable autonomy and tool use, expanding what is possible for a model. For researchers seeking deeper understanding of these components, **Table 1** curates resources for both technical *deep dives* and *hands-on exploration*.

| Chapter Section | Deep Dive References | Recommended Resources |
|---|---|---|
| **1. Pre-training Data** | Jurafsky & Martin (2025) §7.5.2<br>Soldaini et al. (2024) | *Training data visualization*<br>(see footnote 9) |
| **2. Tokenization & Embeddings** | Jurafsky & Martin (2025) §2<br>Voita's NLP Course | *Tokenizer Playground*<br>*Embedding visualizations*<br>(see footnotes 14, 17, 18) |

---

[57] Researchers suggest to rather see LLMs as **cultural and social technologies** (Farrell et al., 2025). Since they have been trained on an enormous amount of text, their real power relies on operating as aggregator and summarizer of such vast knowledge (Veselovsky et al., 2025).

| Chapter Section | Deep Dive References | Recommended Resources |
|---|---|---|
| **3. Transformer Architecture** | Jurafsky & Martin (2025) §7, §8<br>Voita's NLP Course | *Attention visualization tools*<br>*Interactive transformer demos*<br>(see footnotes 23, 25, 26) |
| **4. Probabilistic Generation** | Jurafsky & Martin (2025) §7, §10<br>Voita's NLP Course | *Decoding visualizers*<br>(see footnote 38) |
| **5. Alignment (SFT & RLHF)** | Jurafsky & Martin (2025) §7, §9 | *Fine-tuning platforms*<br>*Community models*<br>(see footnotes 42, 45) |
| **6. Reasoning & Agentic Capabilities** | DeepSeek-AI et al., 2025)<br>Hou et al. (2025)<br>Schick et al. (2023) | *AI Agents course*<br>*MCP server databases*<br>(see footnotes 53, 56) |

*Table 1. Curated Resources for Each LLM Component. Main references for technical depth and hands-on exploration. Specific URLs and detailed descriptions are provided in the corresponding footnotes throughout the chapter.*

The ultimate goal of this chapter is to equip researchers to critically evaluate whether and how LLMs fit their specific research needs, aligning with broader calls for algorithmic and AI literacy in research contexts (Hosseini et al., 2025). Each section has connected technical foundations to practical implications—the **affordances** and **limitations** of using LLMs in research. **Table 2** synthesizes these insights, providing a quick-reference guide that maps each component to its research implications.

| LLM Component | Affordances | Limitations |
|---|---|---|
| **Pre-training Data** | Broad knowledge base<br>Zero-shot capabilities on multiple tasks<br>Transparency in open models | Bounded by training patterns<br>Knowledge cutoff<br>Representational bias<br>Generalization problems |
| **Tokens & Embeddings** | Semantic text operations<br>Robustness to textual variation<br>Processing at scale | Degrading performance on rare/specialized terms<br>Multilingual performance gaps<br>Token-level biases |
| **Attention & Transformer** | Select relevant context<br>Multi-layer inference<br>Process long documents in one pass<br>Efficient deployment techniques | Context window limit<br>Position bias<br>Performance-efficiency trade-offs |
| **Probabilistic Generation** | Coherent long text production<br>Prompt-guided generation<br>Control over output diversity<br>Direct probability analysis | Hallucination and factual errors<br>Lack of logical consistency<br>Bias toward average outputs<br>Stochastic output variability |
| **Alignment** | Follow instructions in natural language<br>Structured output generation | Systematic value biases<br>Sycophantic behavior<br>Overly cautious refusals |

| LLM Component | Affordances | Limitations |
|---|---|---|
| | Accessible customization (e.g., via parameter-efficient methods) Safety and reduced toxicity | |
| **Reasoning & Agentic Abilities** | Enhanced "reasoning" capability Access to external information Autonomous task execution Reusable and shareable tools (via MCP) | Reduced researcher control Higher computational expenses Best practices still being established |

*Table 2. Summary of Six Key Components: Affordances and Limitations for Research. This table provides a condensed reference to the detailed discussions in each section of the chapter.*

*Reason Critically About LLMs*

Throughout this chapter, we have prioritized analytical reasoning over prescriptive guidance. Rather than providing answers for every potential research application, we demonstrate how to *reason critically about LLMs*—identifying which dimensions matter most, systematically evaluating affordances, anticipating limitations, and making informed decisions about deployment. **Box 7** demonstrates this reasoning process in action, walking through an increasingly common application: using LLMs to simulate social media dynamics (and human behavior) in order to explore different recommendation algorithms or moderation policies.

We opened caught between hype and skepticism. We close with something more valuable: the ability to reason through these systems critically. Whether you encounter LLMs in the next chapter on prompting techniques, in discussions of RAG architectures, or in your own research design decisions, you now have the foundational understanding to ask the right questions.

*Box 7. Framework Into Action: LLM-Based Social Media Simulations.*

**The Case Study:** Social media platforms face growing pressure to address polarization and toxicity (Avalle et al., 2024), yet testing interventions like new recommendation algorithms, content moderation policies, or design changes remains challenging. Real-world experiments risk user harm and face ethical, legal, and practical constraints. What if we could replicate social media dynamics in silico and test policies before deployment? This is the promise of LLM-based simulation of social media dynamics. In this approach, researchers use natural language to define both who agents are (very simply put: "behave as are a 35-year-old conservative from Texas") and the possible actions they can do in a given context (write a post, react to another-one etc). Park et al.'s work on *Generative Agents* (Park et al., 2023) was among the first to demonstrate that LLM-based agents could behave in surprisingly realistic ways in simulated environments, using natural language as the medium through which agents both "reason" and "interact." Since then, both commercial ventures (e.g., Artificial Societies, Chirper.ai) and researchers (Gu et al., 2025; Larooij & Törnberg, 2025; Nudo et al., 2025; Orlando et al., 2025; Törnberg et al., 2023; Yang et al., 2025) have built simulated social media platforms to test algorithmic interventions. The appeal is clear: if validated, such simulations could rapidly explore way many scenarios impossible to test otherwise with human subjects.

**The Affordances:** With our framework in hand, we can see exactly why LLM-based simulation looks promising. Consider what each component contributes. Because we know **pre-training** drew on diverse web text, including social media itself, models should already have familiarity with online communication patterns. The transformer architecture's **large context windows** let agents track extended interaction threads, potentially generating responses that account for accumulated conversation history (full thread) (Gu et al., 2025; Nudo et al., 2025). Similarly, understanding the **causal generation** of LLMs reveals why conditioning on a persona's posting history should naturally produce the kinds of posts, replies, and reactions that person would most likely generate; the tokens with highest probability given that context (Chuang, Nirunwiroj, et al., 2024). Thanks to instruction-following capabilities models can be even more explicitly directed through structured prompts that combine persona specifications with behavioral guidelines (Chuang, Goyal, et al., 2024; Larooij & Törnberg, 2025). Researchers can **fine-tune base models** to better capture specific populations or communication styles (Huang et al., 2024), overcoming the overly cautious behavior that alignment produces. Finally, **agentic frameworks** open even richer possibilities. By equipping LLMs with **tools** to scroll feeds, search for relevant posts, or access their own past interactions and generated content, combined with prompts that encourage reflection before deciding whether to engage or stay silent, simulations can achieve far greater realism (Vezhnevets et al., 2023; Yang et al., 2025). In principle, all the building blocks are there.

**The Limitations:** Yet that same framework reveals fundamental tensions. Start with what actually happens in practice. When researchers simulate social media discussions using aligned LLMs, agents often converge toward consensus-oriented outcomes rather than maintaining the conflict and polarization these platforms are known for (Chuang, Goyal, et al., 2024; Coppolillo et al., 2025). We can now see clearly that the explanation lies primarily in **alignment** post-training: models trained to be helpful assistants systematically resist being realistic diverse humans. RLHF-tuned models also exhibit **left-leaning political tendencies** (Bang et al., 2024; Santurkar et al., 2023) which affect also simulations (Coppolillo et al., 2025; A. Liu et al., 2024). Another limitation stems from LLMs being **probabilistic models**. Because LLMs inherently favors common patterns in **pre-training data**, LLMs ultimately mis portray and flatten identity groups (A. Wang et al., 2025). Finally, architectural constraints impose methodological limits on what's feasible. Because these are **decoder-only models**, everything must fit within the **context window**: persona descriptions, interaction history, environmental details, available actions (Park et al., 2023). As simulations become more realistic (more posts, longer threads, richer contexts) researchers face important **design choices** (Chopra et al., 2024; Yang et al., 2025). For example, research needs to decide whether using **larger models** that can guarantee better performances, but computational costs explode for the population-scale simulations of a social media, or smaller, tailored tuned models.

**Acknowledgment**

This work is supported by ERC grant 101140741 Collective Adaptation, FFG grant 873927 Essencse, and NCN/FWF grant 2023/51/I/HS6/02269 Transcend. Views and opinions expressed here do not necessarily reflect those of the funders.

The idea for this chapter originated during the Workshop on The Use of Generative AI for Science of Science and Higher Education Studies (AISci) in Lugano (February 5–7, 2025) and was further developed through interactions at the Social Science and Language Models workshop at the Weizenbaum Institute in Berlin (April 3–4, 2025). I am grateful for the valuable discussions and feedback from participants at both events.

I thank my supervisors Mirta Galesic, Henrik Olsson, and my colleague Victor Møller Poulsen for their thoughtful feedback on drafts of this chapter.

Finally, I acknowledge the use of large language models (Claude Sonnet 4.5) to assist with text editing and rephrasing. All substantive content, visualization and technical explanations are my own work.

# References

Agarwal, A., Mittal, K., Doyle, A., Sridhar, P., Wan, Z., Doughty, J. A., Savelka, J., & Sakr, M. (2024). Understanding the Role of Temperature in Diverse Question Generation by GPT-4. *Proceedings of the 55th ACM Technical Symposium on Computer Science Education V. 2*, 1550–1551. https://doi.org/10.1145/3626253.3635608

Alchokr, R., Borkar, M., Thotadarya, S., Saake, G., & Leich, T. (2023). Supporting systematic literature reviews using deep-learning-based language models. *Proceedings of the 1st International Workshop on Natural Language-Based Software Engineering*, 67–74. https://doi.org/10.1145/3528588.3528658

Allal, L. B., Lozhkov, A., Bakouch, E., Blázquez, G. M., Penedo, G., Tunstall, L., Marafioti, A., Kydlíček, H., Lajarín, A. P., Srivastav, V., Lochner, J., Fahlgren, C., Nguyen, X.-S., Fourrier, C., Burtenshaw, B., Larcher, H., Zhao, H., Zakka, C., Morlon, M., … Wolf, T. (2025). *SmolLM2: When Smol Goes Big -- Data-Centric Training of a Small Language Model* (No. arXiv:2502.02737). arXiv. https://doi.org/10.48550/arXiv.2502.02737

Allen, C., & Hospedales, T. (2019). *Analogies Explained: Towards Understanding Word Embeddings* (No. arXiv:1901.09813). arXiv. https://doi.org/10.48550/arXiv.1901.09813

Amor, M. B., Granitzer, M., & Mitrović, J. (2024). Technical Report: Impact of Position Bias on Language Models in Token Classification. *Proceedings of the 39th ACM/SIGAPP Symposium on Applied Computing*, 741–745. https://doi.org/10.1145/3605098.3636126

Argyle, L. P., Busby, E. C., Fulda, N., Gubler, J., Rytting, C., & Wingate, D. (2023). Out of One, Many: Using Language Models to Simulate Human Samples. *Political Analysis*, *31*(3), 337–351. https://doi.org/10.1017/pan.2023.2

Arora, A., Kaffee, L., & Augenstein, I. (2023). Probing Pre-Trained Language Models for Cross-Cultural Differences in Values. In S. Dev, V. Prabhakaran, D. I. Adelani, D. Hovy, & L.

Benotti (Eds.), *Proceedings of the First Workshop on Cross-Cultural Considerations in NLP (C3NLP)* (pp. 114–130). Association for Computational Linguistics. https://doi.org/10.18653/v1/2023.c3nlp-1.12

Askell, A., Bai, Y., Chen, A., Drain, D., Ganguli, D., Henighan, T., Jones, A., Joseph, N., Mann, B., DasSarma, N., Elhage, N., Hatfield-Dodds, Z., Hernandez, D., Kernion, J., Ndousse, K., Olsson, C., Amodei, D., Brown, T., Clark, J., … Kaplan, J. (2021). *A General Language Assistant as a Laboratory for Alignment* (No. arXiv:2112.00861). arXiv. https://doi.org/10.48550/arXiv.2112.00861

Atari, M., Xue, M. J., Park, P. S., Blasi, D. E., & Henrich, J. (2023). *Which Humans?* https://doi.org/10.31234/osf.io/5b26t

Avalle, M., Di Marco, N., Etta, G., Sangiorgio, E., Alipour, S., Bonetti, A., Alvisi, L., Scala, A., Baronchelli, A., Cinelli, M., & Quattrociocchi, W. (2024). Persistent interaction patterns across social media platforms and over time. *Nature*, *628*(8008), 582–589. https://doi.org/10.1038/s41586-024-07229-y

Bail, C. A. (2024). Can Generative AI improve social science? *Proceedings of the National Academy of Sciences*, *121*(21), e2314021121. https://doi.org/10.1073/pnas.2314021121

Bang, Y., Chen, D., Lee, N., & Fung, P. (2024). *Measuring Political Bias in Large Language Models: What Is Said and How It Is Said* (No. arXiv:2403.18932; Version 1). arXiv. https://doi.org/10.48550/arXiv.2403.18932

Binz, M., Akata, E., Bethge, M., Brändle, F., Callaway, F., Coda-Forno, J., Dayan, P., Demircan, C., Eckstein, M. K., Éltető, N., Griffiths, T. L., Haridi, S., Jagadish, A. K., Ji-An, L., Kipnis, A., Kumar, S., Ludwig, T., Mathony, M., Mattar, M., … Schulz, E. (2025). A foundation model to predict and capture human cognition. *Nature*, *644*(8078), 1002–1009. https://doi.org/10.1038/s41586-025-09215-4

Brown, T., Mann, B., Ryder, N., Subbiah, M., Kaplan, J. D., Dhariwal, P., Neelakantan, A., Shyam, P., Sastry, G., Askell, A., Agarwal, S., Herbert-Voss, A., Krueger, G., Henighan, T., Child, R., Ramesh, A., Ziegler, D., Wu, J., Winter, C., … Amodei, D. (2020). Language Models are Few-Shot Learners. *Advances in Neural Information Processing Systems*, *33*, 1877–1901. https://papers.nips.cc/paper/2020/hash/1457c0d6bfcb4967418bfb8ac142f64a-Abstract.html

Cao, Y., Liu, H., Arora, A., Augenstein, I., Röttger, P., & Hershcovich, D. (2025). Specializing Large Language Models to Simulate Survey Response Distributions for Global Populations. In L. Chiruzzo, A. Ritter, & L. Wang (Eds.), *Proceedings of the 2025 Conference of the Nations of the Americas Chapter of the Association for Computational Linguistics: Human Language Technologies (Volume 1: Long Papers)* (pp. 3141–3154). Association for Computational Linguistics. https://doi.org/10.18653/v1/2025.naacl-long.162

Carolan, K., Fennelly, L., & Smeaton, A. F. (2024). A Review of Multi-Modal Large Language and Vision Models (No. arXiv:2404.01322). arXiv. https://doi.org/10.48550/arXiv.2404.01322

Chai, Y., Fang, Y., Peng, Q., & Li, X. (2024). *Tokenization Falling Short: On Subword Robustness in Large Language Models* (No. arXiv:2406.11687). arXiv. https://doi.org/10.48550/arXiv.2406.11687

Cheng, M., Yu, S., Lee, C., Khadpe, P., Ibrahim, L., & Jurafsky, D. (2025). *ELEPHANT: Measuring and understanding social sycophancy in LLMs* (No. arXiv:2505.13995). arXiv. https://doi.org/10.48550/arXiv.2505.13995

Chopra, A., Kumar, S., Giray-Kuru, N., Raskar, R., & Quera-Bofarull, A. (2024). *On the limits of agency in agent-based models* (No. arXiv:2409.10568; Version 3). arXiv. https://doi.org/10.48550/arXiv.2409.10568

Chuang, Y.-S., Goyal, A., Harlalka, N., Suresh, S., Hawkins, R., Yang, S., Shah, D., Hu, J., & Rogers, T. (2024). Simulating Opinion Dynamics with Networks of LLM-based Agents. In K. Duh, H. Gomez, & S. Bethard (Eds.), *Findings of the Association for Computational Linguistics: NAACL 2024* (pp. 3326–3346). Association for Computational Linguistics. https://doi.org/10.18653/v1/2024.findings-naacl.211

Chuang, Y.-S., Nirunwiroj, K., Studdiford, Z., Goyal, A., Frigo, V. V., Yang, S., Shah, D., Hu, J., & Rogers, T. T. (2024). *Beyond Demographics: Aligning Role-playing LLM-based Agents Using Human Belief Networks* (No. arXiv:2406.17232). arXiv. https://doi.org/10.48550/arXiv.2406.17232

Comanici, G., Bieber, E., Schaekermann, M., Pasupat, I., Sachdeva, N., Dhillon, I., Blistein, M., Ram, O., Zhang, D., Rosen, E., Marris, L., Petulla, S., Gaffney, C., Aharoni, A., Lintz, N., Pais, T. C., Jacobsson, H., Szpektor, I., Jiang, N.-J., … Helmholz, W. (2025). *Gemini 2.5: Pushing the Frontier with Advanced Reasoning, Multimodality, Long Context, and Next Generation Agentic Capabilities* (No. arXiv:2507.06261). arXiv. https://doi.org/10.48550/arXiv.2507.06261

Coppolillo, E., Manco, G., & Aiello, L. M. (2025). *Unmasking Conversational Bias in AI Multiagent Systems* (No. arXiv:2501.14844). arXiv. https://doi.org/10.48550/arXiv.2501.14844

Dagdelen, J., Dunn, A., Lee, S., Walker, N., Rosen, A. S., Ceder, G., Persson, K. A., & Jain, A. (2024). Structured information extraction from scientific text with large language models. *Nature Communications*, *15*(1), 1418. https://doi.org/10.1038/s41467-024-45563-x

DeepSeek-AI, Guo, D., Yang, D., Zhang, H., Song, J., Zhang, R., Xu, R., Zhu, Q., Ma, S., Wang, P., Bi, X., Zhang, X., Yu, X., Wu, Y., Wu, Z. F., Gou, Z., Shao, Z., Li, Z., Gao, Z., … Zhang, Z. (2025). *DeepSeek-R1: Incentivizing Reasoning Capability in LLMs via Reinforcement Learning* (No. arXiv:2501.12948). arXiv. https://doi.org/10.48550/arXiv.2501.12948

DeepSeek-AI, Liu, A., Feng, B., Xue, B., Wang, B., Wu, B., Lu, C., Zhao, C., Deng, C., Zhang, C., Ruan, C., Dai, D., Guo, D., Yang, D., Chen, D., Ji, D., Li, E., Lin, F., Dai, F., … Pan, Z. (2025). *DeepSeek-V3 Technical Report* (No. arXiv:2412.19437). arXiv. https://doi.org/10.48550/arXiv.2412.19437

Dettmers, T., Pagnoni, A., Holtzman, A., & Zettlemoyer, L. (2023). QLoRA: Efficient Finetuning of Quantized LLMs. *Advances in Neural Information Processing Systems*, *36*, 10088–10115.

Devlin, J., Chang, M.-W., Lee, K., & Toutanova, K. (2019). BERT: Pre-training of Deep Bidirectional Transformers for Language Understanding. In J. Burstein, C. Doran, & T. Solorio (Eds.), *Proceedings of the 2019 Conference of the North American Chapter of the Association for Computational Linguistics: Human Language Technologies, Volume 1 (Long and Short Papers)* (pp. 4171–4186). Association for Computational Linguistics. https://doi.org/10.18653/v1/N19-1423

Ding, N., Qin, Y., Yang, G., Wei, F., Yang, Z., Su, Y., Hu, S., Chen, Y., Chan, C.-M., Chen, W., Yi, J., Zhao, W., Wang, X., Liu, Z., Zheng, H.-T., Chen, J., Liu, Y., Tang, J., Li, J., & Sun, M. (2023). Parameter-efficient fine-tuning of large-scale pre-trained language models. *Nature Machine Intelligence*, *5*(3), 220–235. https://doi.org/10.1038/s42256-023-00626-4

Dodge, J., Sap, M., Marasović, A., Agnew, W., Ilharco, G., Groeneveld, D., Mitchell, M., & Gardner, M. (2021). *Documenting Large Webtext Corpora: A Case Study on the Colossal Clean Crawled Corpus* (No. arXiv:2104.08758). arXiv. https://doi.org/10.48550/arXiv.2104.08758

Fan, A., Lewis, M., & Dauphin, Y. (2018). Hierarchical Neural Story Generation. In I. Gurevych & Y. Miyao (Eds.), *Proceedings of the 56th Annual Meeting of the Association for Computational Linguistics (Volume 1: Long Papers)* (pp. 889–898). Association for Computational Linguistics. https://doi.org/10.18653/v1/P18-1082

Farrell, H., Gopnik, A., Shalizi, C., & Evans, J. (2025). Large AI models are cultural and social technologies. *Science*, *387*(6739), 1153–1156. https://doi.org/10.1126/science.adt9819

Ferrando, J., & Voita, E. (2024). *Information Flow Routes: Automatically Interpreting Language Models at Scale* (No. arXiv:2403.00824). arXiv. https://doi.org/10.48550/arXiv.2403.00824

Firth, J. (1957). *A Synopsis of Linguistic Theory, 1930-1955*. https://www.semanticscholar.org/paper/A-Synopsis-of-Linguistic-Theory%2C-1930-1955-Firth/88b3959b6f5333e5358eac43970a5fa29b54642c

Fulay, S., Brannon, W., Mohanty, S., Overney, C., Poole-Dayan, E., Roy, D., & Kabbara, J. (2024). On the Relationship between Truth and Political Bias in Language Models. *Proceedings of* *the 2024 Conference on Empirical Methods in Natural Language Processing*, 9004–9018. https://doi.org/10.18653/v1/2024.emnlp-main.508

Gallegos, I. O., Rossi, R. A., Barrow, J., Tanjim, M. M., Kim, S., Dernoncourt, F., Yu, T., Zhang, R., & Ahmed, N. K. (2024). Bias and Fairness in Large Language Models: A Survey. *Computational Linguistics*, *50*(3), 1097–1179. https://doi.org/10.1162/coli_a_00524

Ghosh, B., Hasan, S., Arafat, N. A., & Khan, A. (2025). *Logical Consistency of Large Language Models in Fact-checking* (No. arXiv:2412.16100; Version 2). arXiv. https://doi.org/10.48550/arXiv.2412.16100

Glickman, M., & Zhang, Y. (2024). AI and Generative AI for Research Discovery and Summarization. *Harvard Data Science Review*, *6*(2). https://doi.org/10.1162/99608f92.7f9220ff

Grattafiori, A., Dubey, A., Jauhri, A., Pandey, A., Kadian, A., Al-Dahle, A., Letman, A., Mathur, A., Schelten, A., Vaughan, A., Yang, A., Fan, A., Goyal, A., Hartshorn, A., Yang, A., Mitra, A., Sravankumar, A., Korenev, A., Hinsvark, A., … Speckbacher, T. (2024). *The Llama 3 Herd of Models*. arXiv. https://doi.org/10.48550/arXiv.2407.21783

Gu, C., Luo, L., Zaidi, Z. R., & Karunasekera, S. (2025). *Large Language Model Driven Agents for Simulating Echo Chamber Formation* (No. arXiv:2502.18138). arXiv. https://doi.org/10.48550/arXiv.2502.18138

Guo, X., & Vosoughi, S. (2025). Serial Position Effects of Large Language Models. In W. Che, J. Nabende, E. Shutova, & M. T. Pilehvar (Eds.), *Findings of the Association for Computational Linguistics: ACL 2025* (pp. 927–953). Association for Computational Linguistics. https://doi.org/10.18653/v1/2025.findings-acl.52

Harris, Z. S. (1954). Distributional Structure. WORD, *10*(2–3), 146–162. https://doi.org/10.1080/00437956.1954.11659520

Heaps, H. S. (1978). *Information Retrieval: Computational and Theoretical Aspects*. Academic Press, Inc.

Hinton, G., Vinyals, O., & Dean, J. (2015). *Distilling the Knowledge in a Neural Network* (No. arXiv:1503.02531). arXiv. https://doi.org/10.48550/arXiv.1503.02531

Holtzman, A., Buys, J., Du, L., Forbes, M., & Choi, Y. (2019). The Curious Case of Neural Text Degeneration. *ArXiv*. https://www.semanticscholar.org/paper/The-Curious-Case-of-Neural-Text-Degeneration-Holtzman-Buys/cf4aa38ae31b43fd07abe13b4ffdb265babb7be1

Holtzman, A., Buys, J., Du, L., Forbes, M., & Choi, Y. (2020). *The Curious Case of Neural Text Degeneration* (No. arXiv:1904.09751). arXiv. https://doi.org/10.48550/arXiv.1904.09751

Hosseini, A., Vani, A., Bahdanau, D., Sordoni, A., & Courville, A. (2022). On the Compositional Generalization Gap of In-Context Learning. In J. Bastings, Y. Belinkov, Y. Elazar, D. Hupkes, N. Saphra, & S. Wiegreffe (Eds.), *Proceedings of the Fifth BlackboxNLP Workshop on Analyzing and Interpreting Neural Networks for NLP* (pp. 272–280). Association for Computational Linguistics. https://doi.org/10.18653/v1/2022.blackboxnlp-1.22

Hosseini, M., Murad, M., & Resnik, D. (2025). *Benefits and Risks of Using AI Agents in Research*. OSF. https://doi.org/10.31235/osf.io/x8sgj_v1

Hou, X., Zhao, Y., Wang, S., & Wang, H. (2025). *Model Context Protocol (MCP): Landscape, Security Threats, and Future Research Directions* (No. arXiv:2503.23278). arXiv. https://doi.org/10.48550/arXiv.2503.23278

Hu, E. J., Shen, Y., Wallis, P., Allen-Zhu, Z., Li, Y., Wang, S., Wang, L., & Chen, W. (2021). *LoRA: Low-Rank Adaptation of Large Language Models* (No. arXiv:2106.09685). arXiv. https://doi.org/10.48550/arXiv.2106.09685

Huang, L., Yu, W., Ma, W., Zhong, W., Feng, Z., Wang, H., Chen, Q., Peng, W., Feng, X., Qin, B., & Liu, T. (2025). A Survey on Hallucination in Large Language Models: Principles, Taxonomy, Challenges, and Open Questions. *ACM Transactions on Information Systems*, *43*(2), 1–55. https://doi.org/10.1145/3703155

Huang, Y., Yuan, Z., Zhou, Y., Guo, K., Wang, X., Zhuang, H., Sun, W., Sun, L., Wang, J., Ye, Y., & Zhang, X. (2024). *Social Science Meets LLMs: How Reliable Are Large Language Models in Social Simulations?* (No. arXiv:2410.23426). arXiv. https://doi.org/10.48550/arXiv.2410.23426

Ibrahim, L., Akbulut, C., Elasmar, R., Rastogi, C., Kahng, M., Morris, M. R., McKee, K. R., Rieser, V., Shanahan, M., & Weidinger, L. (2025). *Multi-turn Evaluation of Anthropomorphic Behaviours in Large Language Models* (No. arXiv:2502.07077). arXiv. https://doi.org/10.48550/arXiv.2502.07077

Ibrahim, L., Collins, K. M., Kim, S. S. Y., Reuel, A., Lamparth, M., Feng, K., Ahmad, L., Soni, P., Kattan, A. E., Stein, M., Swaroop, S., Sucholutsky, I., Strait, A., Liao, Q. V., & Bhatt, U. (2025). *Measuring and mitigating overreliance is necessary for building human-compatible AI* (No. arXiv:2509.08010). arXiv. https://doi.org/10.48550/arXiv.2509.08010

Ji, Z., Lee, N., Frieske, R., Yu, T., Su, D., Xu, Y., Ishii, E., Bang, Y. J., Madotto, A., & Fung, P. (2023). Survey of Hallucination in Natural Language Generation. *ACM Comput. Surv.*, *55*(12), 248:1-248:38. https://doi.org/10.1145/3571730

Jiang, Z., Araki, J., Ding, H., & Neubig, G. (2021). How Can We Know When Language Models Know? On the Calibration of Language Models for Question Answering. *Transactions of the Association for Computational Linguistics*, *9*, 962–977. https://doi.org/10.1162/tacl_a_00407

Joos, M. (1950). Description of Language Design. *The Journal of the Acoustical Society of America*, *22*(6), 701–707. https://doi.org/10.1121/1.1906674

Jurafsky, D., & Martin, J. H. (2025). *Speech and Language Processing: An Introduction to Natural Language Processing, Computational Linguistics, and Speech Recognition with Language Models* (3rd ed.). https://web.stanford.edu/~jurafsky/slp3/

Kadavath, S., Conerly, T., Askell, A., Henighan, T., Drain, D., Perez, E., Schiefer, N., Hatfield-Dodds, Z., DasSarma, N., Tran-Johnson, E., Johnston, S., El-Showk, S., Jones, A., Elhage, N., Hume, T., Chen, A., Bai, Y., Bowman, S., Fort, S., … Kaplan, J. (2022). *Language Models (Mostly) Know What They Know* (No. arXiv:2207.05221). arXiv. https://doi.org/10.48550/arXiv.2207.05221

Kanjirangat, V., Samardzic, T., Dolamic, L., & Rinaldi, F. (2025). Tokenization and Representation Biases in Multilingual Models on Dialectal NLP Tasks. In C. Christodoulopoulos, T. Chakraborty, C. Rose, & V. Peng (Eds.), *Proceedings of the 2025 Conference on Empirical Methods in Natural Language Processing* (pp. 24003–24021). Association for Computational Linguistics. https://doi.org/10.18653/v1/2025.emnlp-main.1224

Kaplan, J., McCandlish, S., Henighan, T., Brown, T. B., Chess, B., Child, R., Gray, S., Radford, A., Wu, J., & Amodei, D. (2020). *Scaling Laws for Neural Language Models* (No. arXiv:2001.08361). arXiv. https://doi.org/10.48550/arXiv.2001.08361

Kudo, T., & Richardson, J. (2018). *SentencePiece: A simple and language independent subword tokenizer and detokenizer for Neural Text Processing* (No. arXiv:1808.06226). arXiv. https://doi.org/10.48550/arXiv.1808.06226

Larooij, M., & Törnberg, P. (2025). *Can We Fix Social Media? Testing Prosocial Interventions using Generative Social Simulation* (No. arXiv:2508.03385). arXiv. https://doi.org/10.48550/arXiv.2508.03385

Laurençon, H., Saulnier, L., Wang, T., Akiki, C., Moral, A. V. del, Scao, T. L., Werra, L. V., Mou, C., Ponferrada, E. G., Nguyen, H., Frohberg, J., Šaško, M., Lhoest, Q., McMillan-Major, A., Dupont, G., Biderman, S., Rogers, A., allal, L. B., Toni, F. D., … Jernite, Y. (2023). *The BigScience ROOTS Corpus: A 1.6TB Composite Multilingual Dataset* (No. arXiv:2303.03915). arXiv. https://doi.org/10.48550/arXiv.2303.03915

Lee, H., Phatale, S., Mansoor, H., Mesnard, T., Ferret, J., Lu, K., Bishop, C., Hall, E., Carbune, V., Rastogi, A., & Prakash, S. (2024). *RLAIF vs. RLHF: Scaling Reinforcement Learning from Human Feedback with AI Feedback* (No. arXiv:2309.00267). arXiv. https://doi.org/10.48550/arXiv.2309.00267

Lee, Y., Oh, J. H., Lee, D., Kang, M., & Lee, S. (2025). Prompt engineering in ChatGPT for literature review: Practical guide exemplified with studies on white phosphors. *Scientific Reports*, *15*(1), 15310. https://doi.org/10.1038/s41598-025-99423-9

Lewis, P., Perez, E., Piktus, A., Petroni, F., Karpukhin, V., Goyal, N., Küttler, H., Lewis, M., Yih, W., Rocktäschel, T., Riedel, S., & Kiela, D. (2020). Retrieval-Augmented Generation for Knowledge-Intensive NLP Tasks. *Advances in Neural Information Processing Systems*, *33*, 9459–9474.

https://proceedings.neurips.cc/paper/2020/hash/6b493230205f780e1bc26945df7481e5-Abstract.html

Liang, W., Zhang, Y., Wu, Z., Lepp, H., Ji, W., Zhao, X., Cao, H., Liu, S., He, S., Huang, Z., Yang, D., Potts, C., Manning, C. D., & Zou, J. Y. (2024). *Mapping the Increasing Use of LLMs in Scientific Papers* (No. arXiv:2404.01268). arXiv. https://doi.org/10.48550/arXiv.2404.01268

Lightman, H., Kosaraju, V., Burda, Y., Edwards, H., Baker, B., Lee, T., Leike, J., Schulman, J., Sutskever, I., & Cobbe, K. (2023). *Let's Verify Step by Step* (No. arXiv:2305.20050). arXiv. https://doi.org/10.48550/arXiv.2305.20050

Lin, J., Tang, J., Tang, H., Yang, S., Chen, W.-M., Wang, W.-C., Xiao, G., Dang, X., Gan, C., & Han, S. (2024). AWQ: Activation-aware Weight Quantization for On-Device LLM Compression and Acceleration. *Proceedings of Machine Learning and Systems*, *6*, 87–100.

Lin, Z. (2025). Techniques for supercharging academic writing with generative AI. *Nature Biomedical Engineering*, *9*(4), 426–431. https://doi.org/10.1038/s41551-024-01185-8

Liu, A., Diab, M., & Fried, D. (2024). Evaluating Large Language Model Biases in Persona-Steered Generation. In L.-W. Ku, A. Martins, & V. Srikumar (Eds.), *Findings of the Association for Computational Linguistics: ACL 2024* (pp. 9832–9850). Association for Computational Linguistics. https://doi.org/10.18653/v1/2024.findings-acl.586

Liu, N. F., Lin, K., Hewitt, J., Paranjape, A., Bevilacqua, M., Petroni, F., & Liang, P. (2024). Lost in the Middle: How Language Models Use Long Contexts. *Transactions of the Association for Computational Linguistics*, *12*, 157–173. https://doi.org/10.1162/tacl_a_00638

Liu, Y., Yao, Y., Ton, J.-F., Zhang, X., Guo, R., Cheng, H., Klochkov, Y., Taufiq, M. F., & Li, H. (2024). *Trustworthy LLMs: A Survey and Guideline for Evaluating Large Language*

*Models' Alignment* (No. arXiv:2308.05374). arXiv. https://doi.org/10.48550/arXiv.2308.05374

Liu, Z., Oguz, B., Zhao, C., Chang, E., Stock, P., Mehdad, Y., Shi, Y., Krishnamoorthi, R., & Chandra, V. (2024). LLM-QAT: Data-Free Quantization Aware Training for Large Language Models. In L.-W. Ku, A. Martins, & V. Srikumar (Eds.), *Findings of the Association for Computational Linguistics: ACL 2024* (pp. 467–484). Association for Computational Linguistics. https://doi.org/10.18653/v1/2024.findings-acl.26

Liu, A., Diab, M., & Fried, D. (2024). *Evaluating Large Language Model Biases in Persona-Steered Generation* (No. arXiv:2405.20253). arXiv. https://doi.org/10.48550/arXiv.2405.20253

Liyanage, C. R., Gokani, R., & Mago, V. (2024). GPT-4 as an X data annotator: Unraveling its performance on a stance classification task. *PLOS ONE*, *19*(8), e0307741. https://doi.org/10.1371/journal.pone.0307741

Lo, K., Wang, L. L., Neumann, M., Kinney, R., & Weld, D. (2020). S2ORC: The Semantic Scholar Open Research Corpus. In D. Jurafsky, J. Chai, N. Schluter, & J. Tetreault (Eds.), *Proceedings of the 58th Annual Meeting of the Association for Computational Linguistics* (pp. 4969–4983). Association for Computational Linguistics. https://doi.org/10.18653/v1/2020.acl-main.447

Lu, C., Lu, C., Lange, R. T., Foerster, J., Clune, J., & Ha, D. (2024). *The AI Scientist: Towards Fully Automated Open-Ended Scientific Discovery* (No. arXiv:2408.06292). arXiv. https://doi.org/10.48550/arXiv.2408.06292

Lu, Y., Li, H., Cong, X., Zhang, Z., Wu, Y., Lin, Y., Liu, Z., Liu, F., & Sun, M. (2025). Learning to Generate Structured Output with Schema Reinforcement Learning. In W. Che, J.

Nabende, E. Shutova, & M. T. Pilehvar (Eds.), *Proceedings of the 63rd Annual Meeting of the Association for Computational Linguistics (Volume 1: Long Papers)* (pp. 4905–4918). Association for Computational Linguistics. https://doi.org/10.18653/v1/2025.acl-long.243

Luo, Z., Yang, Z., Xu, Z., Yang, W., & Du, X. (2025). *LLM4SR: A Survey on Large Language Models for Scientific Research* (No. arXiv:2501.04306). arXiv. https://doi.org/10.48550/arXiv.2501.04306

Maeda, T., & Quan-Haase, A. (2024). When Human-AI Interactions Become Parasocial: Agency and Anthropomorphism in Affective Design. *The 2024 ACM Conference on Fairness, Accountability, and Transparency*, 1068–1077. https://doi.org/10.1145/3630106.3658956

Martinez, R. D., Goriely, Z., Caines, A., Buttery, P., & Beinborn, L. (2024). *Mitigating Frequency Bias and Anisotropy in Language Model Pre-Training with Syntactic Smoothing* (No. arXiv:2410.11462). arXiv. https://doi.org/10.48550/arXiv.2410.11462

Mialon, G., Dessì, R., Lomeli, M., Nalmpantis, C., Pasunuru, R., Raileanu, R., Rozière, B., Schick, T., Dwivedi-Yu, J., Celikyilmaz, A., Grave, E., LeCun, Y., & Scialom, T. (2023). *Augmented Language Models: A Survey* (No. arXiv:2302.07842). arXiv. https://doi.org/10.48550/arXiv.2302.07842

Miao, J., Davis, J. R., Pritchard, J. K., & Zou, J. (2025). *Paper2Agent: Reimagining Research Papers As Interactive and Reliable AI Agents* (No. arXiv:2509.06917). arXiv. https://doi.org/10.48550/arXiv.2509.06917

Miao, J., Davis, J. R., Zhang, Y., Pritchard, J. K., & Zou, J. (2025). *Paper2Agent: Reimagining Research Papers As Interactive and Reliable AI Agents* (No. arXiv:2509.06917). arXiv. https://doi.org/10.48550/arXiv.2509.06917

Mikolov, T., Chen, K., Corrado, G., & Dean, J. (2013). *Efficient Estimation of Word Representations in Vector Space* (No. arXiv:1301.3781). arXiv. https://doi.org/10.48550/arXiv.1301.3781

Mikolov, T., Sutskever, I., Chen, K., Corrado, G. S., & Dean, J. (2013). Distributed Representations of Words and Phrases and their Compositionality. *Advances in Neural Information Processing Systems*, *26*. https://proceedings.neurips.cc/paper_files/paper/2013/hash/9aa42b31882ec039965f3c492 3ce901b-Abstract.html

Mousavi, S. M., Alghisi, S., & Riccardi, G. (2025). *LLMs as Repositories of Factual Knowledge: Limitations and Solutions* (No. arXiv:2501.12774). arXiv. https://doi.org/10.48550/arXiv.2501.12774

Novikova, J., Anderson, C., Blili-Hamelin, B., Rosati, D., & Majumdar, S. (2025). *Consistency in Language Models: Current Landscape, Challenges, and Future Directions* (No. arXiv:2505.00268; Version 2). arXiv. https://doi.org/10.48550/arXiv.2505.00268

Nudo, J., Pandolfo, M. E., Loru, E., Samory, M., Cinelli, M., & Quattrociocchi, W. (2025). *Generative Exaggeration in LLM Social Agents: Consistency, Bias, and Toxicity* (No. arXiv:2507.00657). arXiv. https://doi.org/10.48550/arXiv.2507.00657

OLMo, T., Walsh, P., Soldaini, L., Groeneveld, D., Lo, K., Arora, S., Bhagia, A., Gu, Y., Huang, S., Jordan, M., Lambert, N., Schwenk, D., Tafjord, O., Anderson, T., Atkinson, D., Brahman, F., Clark, C., Dasigi, P., Dziri, N., … Hajishirzi, H. (2025). *2 OLMo 2 Furious* (No. arXiv:2501.00656). arXiv. https://doi.org/10.48550/arXiv.2501.00656

OpenAI, Achiam, J., Adler, S., Agarwal, S., Ahmad, L., Akkaya, I., Aleman, F. L., Almeida, D., Altenschmidt, J., Altman, S., Anadkat, S., Avila, R., Babuschkin, I., Balaji, S., Balcom, V.,

Baltescu, P., Bao, H., Bavarian, M., Belgum, J., … Zoph, B. (2024). *GPT-4 Technical Report* (No. arXiv:2303.08774). arXiv. https://doi.org/10.48550/arXiv.2303.08774

Orlando, G. M., Gatta, V. L., Russo, D., & Moscato, V. (2025). *Can Generative Agent-Based Modeling Replicate the Friendship Paradox in Social Media Simulations?* (No. arXiv:2502.05919). arXiv. https://doi.org/10.48550/arXiv.2502.05919

Osama, R., El-Makky, N., & Torki, M. (2019). Question Answering Using Hierarchical Attention on Top of BERT Features. In A. Fisch, A. Talmor, R. Jia, M. Seo, E. Choi, & D. Chen (Eds.), *Proceedings of the 2nd Workshop on Machine Reading for Question Answering* (pp. 191–195). Association for Computational Linguistics. https://doi.org/10.18653/v1/D19-5825

Ouyang, L., Wu, J., Jiang, X., Almeida, D., Wainwright, C., Mishkin, P., Zhang, C., Agarwal, S., Slama, K., Ray, A., Schulman, J., Hilton, J., Kelton, F., Miller, L., Simens, M., Askell, A., Welinder, P., Christiano, P. F., Leike, J., & Lowe, R. (2022). Training language models to follow instructions with human feedback. *Advances in Neural Information Processing Systems*, *35*, 27730–27744.

Park, J. S., O'Brien, J. C., Cai, C. J., Morris, M. R., Liang, P., & Bernstein, M. S. (2023). *Generative Agents: Interactive Simulacra of Human Behavior* (No. arXiv:2304.03442). arXiv. https://doi.org/10.48550/arXiv.2304.03442

Peng, H., Ke, Q., Budak, C., Romero, D. M., & Ahn, Y.-Y. (2021). Neural embeddings of scholarly periodicals reveal complex disciplinary organizations. *Science Advances*, *7*(17), eabb9004. https://doi.org/10.1126/sciadv.abb9004

Pennington, J., Socher, R., & Manning, C. (2014). GloVe: Global Vectors for Word Representation. In A. Moschitti, B. Pang, & W. Daelemans (Eds.), *Proceedings of the 2014 Conference on*

*Empirical Methods in Natural Language Processing (EMNLP)* (pp. 1532–1543). Association for Computational Linguistics. https://doi.org/10.3115/v1/D14-1162

Petrov, A., La Malfa, E., Torr, P., & Bibi, A. (2023). Language Model Tokenizers Introduce Unfairness Between Languages. *Advances in Neural Information Processing Systems*, *36*, 36963–36990.

Pieuchon, N. A. de, Daoud, A., Jerzak, C. T., Johansson, M., & Johansson, R. (2025). *Benchmarking Debiasing Methods for LLM-based Parameter Estimates* (No. arXiv:2506.09627). arXiv. https://doi.org/10.48550/arXiv.2506.09627

Plaat, A., Duijn, M. van, Stein, N. van, Preuss, M., Putten, P. van der, & Batenburg, K. J. (2025). *Agentic Large Language Models, a survey* (No. arXiv:2503.23037). arXiv. https://doi.org/10.48550/arXiv.2503.23037

Radford, A., Wu, J., Child, R., Luan, D., Amodei, D., & Sutskever, I. (2019). *Language Models are Unsupervised Multitask Learners*. https://www.semanticscholar.org/paper/Language-Models-are-Unsupervised-Multitask-Learners-Radford-Wu/9405cc0d6169988371b2755e573cc28650d14dfe

Rafailov, R., Sharma, A., Mitchell, E., Manning, C. D., Ermon, S., & Finn, C. (2023). Direct Preference Optimization: Your Language Model is Secretly a Reward Model. *Advances in Neural Information Processing Systems*, *36*, 53728–53741.

Raffel, C., Shazeer, N., Roberts, A., Lee, K., Narang, S., Matena, M., Zhou, Y., Li, W., & Liu, P. J. (2023). *Exploring the Limits of Transfer Learning with a Unified Text-to-Text Transformer* (No. arXiv:1910.10683). arXiv. https://doi.org/10.48550/arXiv.1910.10683

Rawte, V., Chakraborty, S., Pathak, A., Sarkar, A., Tonmoy, S. M. T. I., Chadha, A., Sheth, A., & Das, A. (2023). The Troubling Emergence of Hallucination in Large Language Models—

An Extensive Definition, Quantification, and Prescriptive Remediations. In H. Bouamor, J. Pino, & K. Bali (Eds.), *Proceedings of the 2023 Conference on Empirical Methods in Natural Language Processing* (pp. 2541–2573). Association for Computational Linguistics. https://doi.org/10.18653/v1/2023.emnlp-main.155

Reid, M., Savinov, N., Teplyashin, D., Lepikhin, D., Lillicrap, T., Alayrac, J.-B., Soricut, R., Lazaridou, A., Firat, O., Schrittwieser, J., Antonoglou, I., Anil, R., Borgeaud, S., Dai, A. M., Millican, K., Dyer, E., Glaese, M., Sottiaux, T., Lee, B., … Chronopoulou, A. (2024). Gemini 1.5: Unlocking multimodal understanding across millions of tokens of context. *ArXiv*. https://www.semanticscholar.org/paper/Gemini-1.5%3A-Unlocking-multimodal-understanding-of-Reid-Savinov/c811bedbe8f4c21d0cba9f9175f7c2eb203284a7

Ross, A. S. C., & Herdan, G. (1960). Type-Token Mathematics: A Textbook of Mathematical Linguistics. *Journal of the Royal Statistical Society. Series A (General)*, *123*(3), 341. https://doi.org/10.2307/2342480

Santurkar, S., Durmus, E., Ladhak, F., Lee, C., Liang, P., & Hashimoto, T. (2023). Whose Opinions Do Language Models Reflect? *Proceedings of the 40th International Conference on Machine Learning*, 29971–30004. https://proceedings.mlr.press/v202/santurkar23a.html

Schick, T., Dwivedi-Yu, J., Dessi, R., Raileanu, R., Lomeli, M., Hambro, E., Zettlemoyer, L., Cancedda, N., & Scialom, T. (2023). Toolformer: Language Models Can Teach Themselves to Use Tools. *Advances in Neural Information Processing Systems*, *36*, 68539–68551.

Schulman, J., Wolski, F., Dhariwal, P., Radford, A., & Klimov, O. (2017). *Proximal Policy Optimization Algorithms* (No. arXiv:1707.06347). arXiv. https://doi.org/10.48550/arXiv.1707.06347

Sen, I., Lutz, M., Rogers, E., Garcia, D., & Strohmaier, M. (2025). Missing the Margins: A Systematic Literature Review on the Demographic Representativeness of LLMs. In W. Che, J. Nabende, E. Shutova, & M. T. Pilehvar (Eds.), *Findings of the Association for Computational Linguistics: ACL 2025* (pp. 24263–24289). Association for Computational Linguistics. https://doi.org/10.18653/v1/2025.findings-acl.1246

Sennrich, R., Haddow, B., & Birch, A. (2016a). Neural Machine Translation of Rare Words with Subword Units. In K. Erk & N. A. Smith (Eds.), *Proceedings of the 54th Annual Meeting of the Association for Computational Linguistics (Volume 1: Long Papers)* (pp. 1715–1725). Association for Computational Linguistics. https://doi.org/10.18653/v1/P16-1162

Sennrich, R., Haddow, B., & Birch, A. (2016b). Neural Machine Translation of Rare Words with Subword Units. In K. Erk & N. A. Smith (Eds.), *Proceedings of the 54th Annual Meeting of the Association for Computational Linguistics (Volume 1: Long Papers)* (pp. 1715–1725). Association for Computational Linguistics. https://doi.org/10.18653/v1/P16-1162

Sharma, M., Tong, M., Korbak, T., Duvenaud, D., Askell, A., Bowman, S. R., Cheng, N., Durmus, E., Hatfield-Dodds, Z., Johnston, S. R., Kravec, S., Maxwell, T., McCandlish, S., Ndousse, K., Rausch, O., Schiefer, N., Yan, D., Zhang, M., & Perez, E. (2025). *Towards Understanding Sycophancy in Language Models* (No. arXiv:2310.13548). arXiv. https://doi.org/10.48550/arXiv.2310.13548

Shibayama, S., Yin, D., & Matsumoto, K. (2021). Measuring novelty in science with word embedding. *PLOS ONE*, *16*(7), e0254034. https://doi.org/10.1371/journal.pone.0254034

Soldaini, L., Kinney, R., Bhagia, A., Schwenk, D., Atkinson, D., Authur, R., Bogin, B., Chandu, K., Dumas, J., Elazar, Y., Hofmann, V., Jha, A. H., Kumar, S., Lucy, L., Lyu, X., Lambert, N., Magnusson, I., Morrison, J., Muennighoff, N., … Lo, K. (2024). *Dolma: An Open*

*Corpus of Three Trillion Tokens for Language Model Pretraining Research* (No. arXiv:2402.00159). arXiv. https://doi.org/10.48550/arXiv.2402.00159

Suzgun, M., Gur, T., Bianchi, F., Ho, D. E., Icard, T., Jurafsky, D., & Zou, J. (2025). Language models cannot reliably distinguish belief from knowledge and fact. *Nature Machine Intelligence*, *7*(11), 1780–1790. https://doi.org/10.1038/s42256-025-01113-8

Suzgun, M., Scales, N., Schärli, N., Gehrmann, S., Tay, Y., Chung, H. W., Chowdhery, A., Le, Q., Chi, E., Zhou, D., & Wei, J. (2023). Challenging BIG-Bench Tasks and Whether Chain-of-Thought Can Solve Them. In A. Rogers, J. Boyd-Graber, & N. Okazaki (Eds.), *Findings of the Association for Computational Linguistics: ACL 2023* (pp. 13003–13051). Association for Computational Linguistics. https://doi.org/10.18653/v1/2023.findings-acl.824

Tang, X., Duan, X., & Cai, Z. (2025). Large Language Models for Automated Literature Review: An Evaluation of Reference Generation, Abstract Writing, and Review Composition. In C. Christodoulopoulos, T. Chakraborty, C. Rose, & V. Peng (Eds.), *Proceedings of the 2025 Conference on Empirical Methods in Natural Language Processing* (pp. 1602–1617). Association for Computational Linguistics. https://doi.org/10.18653/v1/2025.emnlp-main.83

Tao, Y., Viberg, O., Baker, R. S., & Kizilcec, R. F. (2024). Cultural bias and cultural alignment of large language models. *PNAS Nexus*, *3*(9), pgae346. https://doi.org/10.1093/pnasnexus/pgae346

Team, G., Kamath, A., Ferret, J., Pathak, S., Vieillard, N., Merhej, R., Perrin, S., Matejovicova, T., Ramé, A., Rivière, M., Rouillard, L., Mesnard, T., Cideron, G., Grill, J., Ramos, S., Yvinec,

E., Casbon, M., Pot, E., Penchev, I., … Hussenot, L. (2025). *Gemma 3 Technical Report* (No. arXiv:2503.19786). arXiv. https://doi.org/10.48550/arXiv.2503.19786

Tenney, I., Xia, P., Chen, B., Wang, A., Poliak, A., McCoy, R. T., Kim, N., Durme, B. V., Bowman, S. R., Das, D., & Pavlick, E. (2019). *What do you learn from context? Probing for sentence structure in contextualized word representations* (No. arXiv:1905.06316). arXiv. https://doi.org/10.48550/arXiv.1905.06316

Törnberg, P. (2023). *ChatGPT-4 Outperforms Experts and Crowd Workers in Annotating Political Twitter Messages with Zero-Shot Learning* (No. arXiv:2304.06588). arXiv. https://doi.org/10.48550/arXiv.2304.06588

Törnberg, P. (2024). Best Practices for Text Annotation with Large Language Models. *Sociologica*, *18*(2), 67–85. https://doi.org/10.6092/issn.1971-8853/19461

Törnberg, P., Valeeva, D., Uitermark, J., & Bail, C. (2023). *Simulating Social Media Using Large Language Models to Evaluate Alternative News Feed Algorithms* (No. arXiv:2310.05984). arXiv. https://doi.org/10.48550/arXiv.2310.05984

Vaswani, A., Shazeer, N., Parmar, N., Uszkoreit, J., Jones, L., Gomez, A. N., Kaiser, Ł. ukasz, & Polosukhin, I. (2017). Attention is All you Need. *Advances in Neural Information Processing Systems*, *30*. https://proceedings.neurips.cc/paper_files/paper/2017/hash/3f5ee243547dee91fbd053c1c4a845aa-Abstract.html

Veselovsky, V., Argin, B., Stroebl, B., Wendler, C., West, R., Evans, J., Griffiths, T. L., & Narayanan, A. (2025). *Localized Cultural Knowledge is Conserved and Controllable in Large Language Models* (No. arXiv:2504.10191). arXiv. https://doi.org/10.48550/arXiv.2504.10191

Vezhnevets, A. S., Agapiou, J. P., Aharon, A., Ziv, R., Matyas, J., Duéñez-Guzmán, E. A., Cunningham, W. A., Osindero, S., Karmon, D., & Leibo, J. Z. (2023). *Generative agent-based modeling with actions grounded in physical, social, or digital space using Concordia* (No. arXiv:2312.03664). arXiv. https://doi.org/10.48550/arXiv.2312.03664

Voita, E., Talbot, D., Moiseev, F., Sennrich, R., & Titov, I. (2019). Analyzing Multi-Head Self-Attention: Specialized Heads Do the Heavy Lifting, the Rest Can Be Pruned. In A. Korhonen, D. Traum, & L. Màrquez (Eds.), *Proceedings of the 57th Annual Meeting of the Association for Computational Linguistics* (pp. 5797–5808). Association for Computational Linguistics. https://doi.org/10.18653/v1/P19-1580

Voita, E. (2020). NLP Course For You. https://lena-voita.github.io/nlp_course.html (last accessed 23.12.2025).

Wan, Z., Wang, X., Liu, C., Alam, S., Zheng, Y., Liu, J., Qu, Z., Yan, S., Zhu, Y., Zhang, Q., Chowdhury, M., & Zhang, M. (2024). *Efficient Large Language Models: A Survey* (No. arXiv:2312.03863). arXiv. https://doi.org/10.48550/arXiv.2312.03863

Wang, A., Morgenstern, J., & Dickerson, J. P. (2025). Large language models that replace human participants can harmfully misportray and flatten identity groups. *Nature Machine Intelligence*, *7*(3), 400–411. https://doi.org/10.1038/s42256-025-00986-z

Wang, T., Roberts, A., Hesslow, D., Scao, T. L., Chung, H. W., Beltagy, I., Launay, J., & Raffel, C. (2022). *What Language Model Architecture and Pretraining Objective Work Best for Zero-Shot Generalization?* (No. arXiv:2204.05832). arXiv. https://doi.org/10.48550/arXiv.2204.05832

Wang, Y., Wang, M., Manzoor, M. A., Liu, F., Georgiev, G. N., Das, R. J., & Nakov, P. (2024). Factuality of Large Language Models: A Survey. In Y. Al-Onaizan, M. Bansal, & Y.-N.

Chen (Eds.), *Proceedings of the 2024 Conference on Empirical Methods in Natural Language Processing* (pp. 19519–19529). Association for Computational Linguistics. https://doi.org/10.18653/v1/2024.emnlp-main.1088

Wei, J., Tay, Y., Bommasani, R., Raffel, C., Zoph, B., Borgeaud, S., Yogatama, D., Bosma, M., Zhou, D., Metzler, D., Chi, E. H., Hashimoto, T., Vinyals, O., Liang, P., Dean, J., & Fedus, W. (2022). *Emergent Abilities of Large Language Models* (No. arXiv:2206.07682). arXiv. https://doi.org/10.48550/arXiv.2206.07682

Wei, J., Wang, X., Schuurmans, D., Bosma, M., Ichter, B., Xia, F., Chi, E., Le, Q., & Zhou, D. (2023). *Chain-of-Thought Prompting Elicits Reasoning in Large Language Models* (No. arXiv:2201.11903). arXiv. https://doi.org/10.48550/arXiv.2201.11903

Weidinger, L., Mellor, J., Rauh, M., Griffin, C., Uesato, J., Huang, P.-S., Cheng, M., Glaese, M., Balle, B., Kasirzadeh, A., Kenton, Z., Brown, S., Hawkins, W., Stepleton, T., Biles, C., Birhane, A., Haas, J., Rimell, L., Hendricks, L. A., … Gabriel, I. (2021). *Ethical and social risks of harm from Language Models* (No. arXiv:2112.04359). arXiv. https://doi.org/10.48550/arXiv.2112.04359

Wuttke, A., Aßenmacher, M., Klamm, C., Lang, M. M., Würschinger, Q., & Kreuter, F. (2025). AI Conversational Interviewing: Transforming Surveys with LLMs as Adaptive Interviewers. In A. Kazantseva, S. Szpakowicz, S. Degaetano-Ortlieb, Y. Bizzoni, & J. Pagel (Eds.), *Proceedings of the 9th Joint SIGHUM Workshop on Computational Linguistics for Cultural Heritage, Social Sciences, Humanities and Literature (LaTeCH-CLfL 2025)* (pp. 179–204). Association for Computational Linguistics. https://doi.org/10.18653/v1/2025.latechclfl-1.17

Yang, A., Li, A., Yang, B., Zhang, B., Hui, B., Zheng, B., Yu, B., Gao, C., Huang, C., Lv, C., Zheng, C., Liu, D., Zhou, F., Huang, F., Hu, F., Ge, H., Wei, H., Lin, H., Tang, J., … Qiu, Z. (2025). *Qwen3 Technical Report* (No. arXiv:2505.09388). arXiv. https://doi.org/10.48550/arXiv.2505.09388

Yang, Z., Du, X., Li, J., Zheng, J., Poria, S., & Cambria, E. (2024). Large Language Models for Automated Open-domain Scientific Hypotheses Discovery. In L.-W. Ku, A. Martins, & V. Srikumar (Eds.), *Findings of the Association for Computational Linguistics: ACL 2024* (pp. 13545–13565). Association for Computational Linguistics. https://doi.org/10.18653/v1/2024.findings-acl.804

Yang, Z., Zhang, Z., Zheng, Z., Jiang, Y., Gan, Z., Wang, Z., Ling, Z., Chen, J., Ma, M., Dong, B., Gupta, P., Hu, S., Yin, Z., Li, G., Jia, X., Wang, L., Ghanem, B., Lu, H., Lu, C., … Shao, J. (2025). *OASIS: Open Agent Social Interaction Simulations with One Million Agents* (No. arXiv:2411.11581). arXiv. https://doi.org/10.48550/arXiv.2411.11581

Yao, S., Zhao, J., Yu, D., Du, N., Shafran, I., Narasimhan, K., & Cao, Y. (2023). *ReAct: Synergizing Reasoning and Acting in Language Models* (No. arXiv:2210.03629). arXiv. https://doi.org/10.48550/arXiv.2210.03629

Zhang, H., Duckworth, D., Ippolito, D., & Neelakantan, A. (2021). Trading Off Diversity and Quality in Natural Language Generation. In A. Belz, S. Agarwal, Y. Graham, E. Reiter, & A. Shimorina (Eds.), *Proceedings of the Workshop on Human Evaluation of NLP Systems (HumEval)* (pp. 25–33). Association for Computational Linguistics. https://aclanthology.org/2021.humeval-1.3/

Zhang, M., Huang, M., Shi, R., Guo, L., Peng, C., Yan, P., Zhou, Y., & Qiu, X. (2024). Calibrating the Confidence of Large Language Models by Eliciting Fidelity. In Y. Al-Onaizan, M.

Bansal, & Y.-N. Chen (Eds.), *Proceedings of the 2024 Conference on Empirical Methods in Natural Language Processing* (pp. 2959–2979). Association for Computational Linguistics. https://doi.org/10.18653/v1/2024.emnlp-main.173

Zhang, X., Chen, Y., Hu, S., Xu, Z., Chen, J., Hao, M., Han, X., Thai, Z., Wang, S., Liu, Z., & Sun, M. (2024). ∞ftyBench: Extending Long Context Evaluation Beyond 100K Tokens. In L.-W. Ku, A. Martins, & V. Srikumar (Eds.), *Proceedings of the 62nd Annual Meeting of the Association for Computational Linguistics (Volume 1: Long Papers)* (pp. 15262–15277). Association for Computational Linguistics. https://doi.org/10.18653/v1/2024.acl-long.814

Zhao, W. X., Zhou, K., Li, J., Tang, T., Wang, X., Hou, Y., Min, Y., Zhang, B., Zhang, J., Dong, Z., Du, Y., Yang, C., Chen, Y., Chen, Z., Jiang, J., Ren, R., Li, Y., Tang, X., Liu, Z., … Wen, J.-R. (2025). *A Survey of Large Language Models* (No. arXiv:2303.18223). arXiv. https://doi.org/10.48550/arXiv.2303.18223

Zhu, X., Li, J., Liu, Y., Ma, C., & Wang, W. (2024). A Survey on Model Compression for Large Language Models. *Transactions of the Association for Computational Linguistics*, *12*, 1556–1577. https://doi.org/10.1162/tacl_a_00704

Ziems, C., Held, W., Shaikh, O., Chen, J., Zhang, Z., & Yang, D. (2024). Can Large Language Models Transform Computational Social Science? *Computational Linguistics*, *50*(1), 237–291. https://doi.org/10.1162/coli_a_00502